\documentclass{article}
\usepackage{ijcai21}

\usepackage{xr-hyper}
\usepackage{times}
\usepackage{soul}
\usepackage{url}
\usepackage[hidelinks]{hyperref}
\usepackage[utf8]{inputenc}
\usepackage[small]{caption}
\usepackage{graphicx}
\usepackage{amsmath}
\usepackage{amsthm}
\usepackage{amssymb}
\usepackage{booktabs}
\usepackage{algorithm}
\usepackage{algorithmic}
\usepackage{pgfplots}
\usepackage{tikz}
\usepackage{subfig}
\usepackage{xcolor}

\usetikzlibrary{shapes.geometric}
\usetikzlibrary{positioning,fit,shapes,shadows,arrows,calc}
\usetikzlibrary{intersections}
\pgfplotsset{compat=1.15}
\pgfdeclarelayer{background}
\pgfdeclarelayer{foreground}
\pgfsetlayers{background,main,foreground}

\usepackage{natbib}

\title{DACBench: A Benchmark Library for Dynamic Algorithm Configuration}

\author{
Theresa Eimer$^1$\and
André Biedenkapp$^2$\and
Maximilian Reimer$^1$\and
Steven Adriaensen$^2$\and\\\vspace*{0.1cm}
Frank Hutter$^{2,3}$\And
Marius Lindauer$^{1}$
\affiliations
$^1$Information Processing Institute (tnt),  Leibniz University Hannover, Germany\\
$^2$Department  of  Computer  Science,  University  of  Freiburg, Germany\\
$^3$Bosch Center for Artificial Intelligence, Renningen, Germany
\emails
\{eimer, reimerm, lindauer\}@tnt.uni-hannover.de,
\{biedenka, adriaens, fh\}@cs.uni-freiburg.de
}

\begin{document}

\maketitle

\begin{abstract}
Dynamic Algorithm Configuration (DAC) aims to dynamically control a target algorithm's hyperparameters in order to improve its performance.
Several theoretical and empirical results have demonstrated the benefits of dynamically controlling hyperparameters in domains like evolutionary computation, AI Planning or deep learning.
Replicating these results, as well as studying new methods for DAC, however, is difficult since existing benchmarks are often specialized and incompatible with the same interfaces.
To facilitate benchmarking and thus research on DAC, 
we propose DACBench, a benchmark library that seeks to collect and standardize existing DAC benchmarks from different AI domains, as well as provide a template for new ones.
For the design of DACBench, we focused on important desiderata, such as (i) flexibility, (ii) reproducibility, (iii) extensibility and (iv) automatic documentation and visualization.
To show the potential, broad applicability and challenges of DAC, we explore how a set of six initial benchmarks compare in several dimensions of difficulty.
\end{abstract}

\section{Introduction}
In the last years, algorithm configuration~\citep{hutter-jair09a,ansotegui-cp09a,hutter-lion11a,lopez-ibanez-orp16} and in particular automated machine learning~\citep{shahriari-ieee16a,hutter-book19a} offered automatic methods optimizing the settings of hyperparameters to improve the performance of algorithms. 
However, practitioners of different communities have already known for a while that static hyperparameter settings do not necessarily yield optimal performance compared to dynamic hyperparameter policies~\citep{senior-icassp13}.
One way of formalizing dynamic adaptations of hyperparameters is dynamic algorithm configuration \citep[DAC;][]{biedenkapp-ecai20}.
DAC showed its promise by outperforming other algorithm configuration approaches, e.g., choosing variants of CMA-ES~\citep{vermetten-gecco19} or dynamically adapting its step-size~\citep{shala-ppsn20}, dynamically switching between heuristics in AI planning~\citep{speck-icaps21}, or learning learning rate schedules for computer vision~\citep{daniel-aaai16}.

These results, however, also revealed a challenge for the further development of DAC. 
Compared to static algorithm configuration~\citep{hutter-jair09a,ansotegui-cp09a,hutter-lion11a,lopez-ibanez-orp16},  applying DAC also requires (i) the definition of a configuration space to search in, (ii) instances to optimize on and (iii) a reward signal defining the quality of hyperparameter settings. 
However, the optimizer and the algorithm to be optimized have to be integrated much closer  in DAC. 
The current state of the algorithm and the reward function, for example, need to be queried by the optimizer on a regular basis and the applied hyperparameter changes need to be communicated to the algorithm. 
Therefore, creating reliable, reusable and easy-to-use DAC benchmarks is often fairly hard with no existing standard thus far.

This disparity between benchmarks in addition to the difficulty in creating new ones presents a barrier of entry to the field. 
Researchers not well versed in both target domain and DAC may not be able to reproduce experiments or understand the way benchmarks are modelled.
This makes it hard for pure domain experts to create a DAC benchmark for their domain, severely limiting the number of future benchmarks we can expect to see.
A lack of standardized benchmarks, in turn, will slow the progress of DAC going forward as there is no reliable way to compare methods on a diverse set of problems.

To close this gap, we propose DACBench, a suite of standardized benchmarks\footnote{The project repository can be found at https://github.com/automl/DACBench}. 
On one hand, we integrate a diverse set of AI algorithms from different domains, such as AI planning, deep learning and evolutionary computation. 
On the other hand, we ensure that all benchmarks can be used with a unified easy-to-use interface, that allows the application of a multitude of different DAC approaches as well as the simple addition of new benchmarks.
This paper details the concepts and ideas of DACBench, as well as insights from the benchmarks themselves. Specifically, our contributions are:
\begin{enumerate}
    \item We propose DACBench, a DAC benchmark suite with a standardized interface and tools to ensure comparability and reproducibility of results;
    \item We discuss desiderata of creating DAC benchmarks and how we took them into account in DACBench;
    \item We propose a diverse set of DAC benchmarks from different domains showing the  breadth of DAC's potential, allowing future research to make strong claims with new DAC methods;
    \item We show that our DAC benchmarks cover different challenges in DAC application and research.
\end{enumerate}

With this, we strive to lower the barrier of entrance into DAC research and enable research that matters.

\section{Related Work}
DAC is a general way to formulate the problem of optimizing the performance of an algorithm by dynamically adapting its hyperparameters, subsuming both algorithm configuration \citep[AC;][]{hutter-aij17a} and per-instance algorithm configuration \citep[PIAC; e.g. ][]{ansotegui-aij16}. 
While AC methods can achieve significant improvements over default configurations PIAC algorithms have demonstrated that searching for a configuration per instance can further improve performance.
In a similar way, DAC can navigate the over time changing search landscape in addition to instance-specific variations.

Theoretically, this has been shown to be optimal for the $(1 + (\lambda, \lambda))$ genetic algorithm \citep{doerr-algo18}, and to enable exponential speedups compared to AC on a family of AI Planning problems \citep{speck-icaps21}.

Empirically, we have seen dynamic hyperparameter schedules outperform static settings in fields like Evolutionary Computation~\citep{shala-ppsn20}, AI Planning~\citep{speck-icaps21} and Deep Learning~\citep{daniel-aaai16}.
In addition, hyperheuristics~\citep{burke-mista09} can also be seen as a form of DAC. In this field, it has been shown that dynamic heuristic selection outperforms static approaches on combinatorial optimization problems like Knapsack or Max-Cut \citep{almutairi-iscis16}.

In the context of machine learning, dynamically adjusting an algorithm's hyperparameters can be seen as a form of learning to learn where the goal is to learn algorithms or algorithm components like loss functions \citep{houthooft-neurips18}, exploration strategies \citep{gupta-neurips18} or completely new algorithms \citep{andrychowicz-neurips16,chen-icml17}. 
While DAC does not attempt to replace algorithm components with learned ones, the hyperparameter values of an algorithm are often instrumental in guiding its progress. In some cases they become part of the algorithm. Dynamic step size adaption in ES using heuristics, for example, is very common, but can be replaced and outperformed by more specific DAC hyperparameter policies \citep{shala-ppsn20}.

In other meta-algorithmic areas, reliable and well engineered benchmark libraries also facilitated research progress, incl. ASLib~\citep{bischl-aij16a}, ACLib~\citep{hutter-lion14a}, tabular NAS benchmarks \citep[e.g.,][]{ying-icml19a} and HPOlib~\citep{eggensperger-bayesopt13}. In particular, DACBench is strongly inspired by HPOlib and OpenAI gym~\citep{gym} which also provide a unified interface to benchmarks. Although the hyflex framework~\citep{burke-mista09} addresses a similar meta-algorithmic problem, in DACBench, we can model more complex problems (i.e., continuous and mixed spaces instead of only categoricals), consider state features of algorithms and cover more AI domains (not only combinatorial problems).

Furthermore DACBench is designed to build upon existing benchmark libraries in target domains by integrating their algorithm implementations.
This includes well-established benchmarks like COCO \citep{hansen-oms2020} or IOHProfiler \citep{IOHprofiler}.

\section{Formal Background on DAC}
DAC aims at improving a target algorithm's performance through dynamic control of its hyperparameter settings \mbox{$\mathbf{\lambda} \in \Lambda$}. To this end, a DAC policy $\pi$ queries state information $s_t \in \mathcal{S}$ of the target algorithm at each time point $t$ to set a hyperparameter configuration: $\pi : \mathcal{S} \to \Lambda$.
Given a starting state $s_0$ of the target algorithm, a maximal number of solving steps $T$, a probability distribution $p$ over a space of problem instances $i \in \mathcal{I}$, and a reward function $r_i: \mathcal{S} \times \Lambda \to \mathbb{R}$ depending on the instance $i$ at hand, the objective is to find a policy maximizing the total return:
\begin{equation}
    \int_{\mathcal{I}} p(i) \sum_{t=0}^T r_i(s_t, \pi(s_t)) \, \mathrm{d}i
\end{equation}

Following \cite{biedenkapp-ecai20}, one way of modelling this task is as a contextual MDP $M_\mathcal{I} = \{ M_i \}_{i \sim \mathcal{I}}$ \citep{hallak-corr15}, consisting of $|\mathcal{I}|$ MDPs. Each $M_i$ represents one target problem instance $i$ with $M_i = (\mathcal{S}, \mathcal{A}, \mathcal{T}_i, r_i)$. This formulation assumes that all $M_i$ share a common state space $\mathcal{S}$, describing all possible algorithm states, as well as a single action space $\mathcal{A}$ choosing from all possible hyperparameter configurations $\Lambda$. 
The transition function $\mathcal{T}_i: \mathcal{S} \times \mathcal{A} \to \mathcal{S}$, corresponding to algorithm behaviour, and reward function~$r_i$, however, vary between instances. 

This formulation allows to apply different configuration approaches on the same problem setting, e.g., algorithm configuration by ignoring all state information $(\pi: \emptyset \to \Lambda)$, per-instance algorithm configuration by only taking the instance into account $(\pi: \mathcal{I} \to \Lambda)$ or a full DAC agent $(\pi: \mathcal{S} \times \mathcal{I} \to \Lambda)$ on the contextual MDP (information about $i\in \mathcal{I}$ is typically directly reflected in $s\in\mathcal{S}$). In view of how close this DAC formulation is to reinforcement learning (RL), in the remainder of the paper we will continue to refer to hyperparameter settings as actions and hyperparameter schedules as policies.
Nevertheless, we consider DAC as a general problem that can be solved in different ways, incl. supervised learning, reinforcement learning or even hand-designed policies, e.g., cosine annealing for learning rate adaption in deep learning~\citep{loshchilov-iclr17a} or CSA for CMA-ES~\citep{chotard-ppsn12}.

\section{DACBench}

\tikzstyle{activity}=[align=center, rectangle, draw=black, rounded corners, text width=8em, fill=white, drop shadow]
\tikzstyle{data}=[align=center, rectangle, draw=black, fill=black!10, text width=8em, drop shadow]
\tikzstyle{internal}=[rectangle, draw=black!60, thick, align=left, text width=8em, minimum height=.75cm, text=black!60]
\tikzstyle{myarrow}=[-latex, thick]

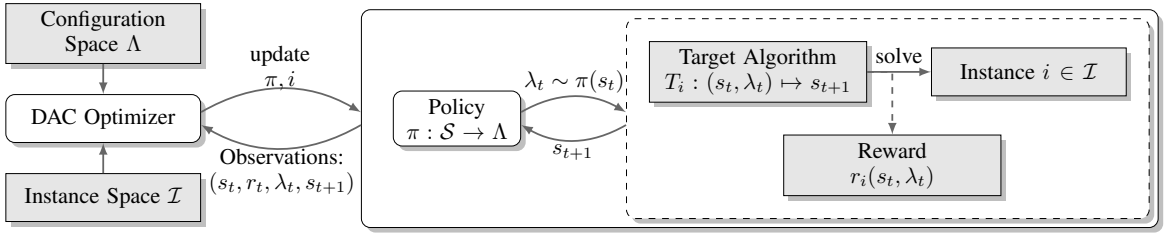
\begin{figure*}[tbp]
\centering
   \scalebox{.85}{
	\begin{tikzpicture}
	
	\node (optimizer) [activity,minimum height=.75cm] {DAC Optimizer};
	\node (space) [data,minimum height=.75cm, above=of optimizer, yshift=-1.5em] {Configuration Space $\Lambda$};
	\node (insts) [data,minimum height=.75cm, below=of optimizer, yshift=1.5em] {Instance Space $\mathcal{I}$};
	\node (policy) [activity,minimum height=.75cm, right=of optimizer, text width=5em, xshift=2.0cm] {Policy $\pi: \mathcal{S} \to \Lambda$};
	
	\node (algo) [data, minimum height=.75cm, right=of policy, text width=9em, yshift=0.73cm, xshift=1cm] {Target Algorithm $T_i: (s_t, \lambda_t) \mapsto s_{t+1}$};
	
	\node (inst) [data, minimum height=.75cm, right=of algo, text width=8em] {Instance $i \in \mathcal{I}$};
	
	\node (reward) [data, minimum height=.75cm, below=of algo, text width=9em, xshift=2.1cm, yshift=0.5cm] {Reward\\ $r_i(s_t,\lambda_t)$};
	
	\begin{pgfonlayer}{background}
	    \node (outer) [activity, inner sep=14pt, fit={(policy) (algo) (inst) (reward)}] {};
        \node (int) [activity, dashed, inner sep=10pt, fit={(algo) (inst) (reward)}] {};
    \end{pgfonlayer}
    
    % inner arrows
    \draw[myarrow,draw=black!60, dashed] (algo.east) -| (reward.north);
    \draw[myarrow,draw=black!60] (algo.east) -- (inst.west) node[above, align=center, pos=0.5]{solve};
    %\draw[myarrow,draw=black!60, dashed] (reward.west) -| (algo.south);

    % mid arrows
    \draw[myarrow,draw=black!60] ($(policy.east)+(0,0.1)$) to [in=150,out=30] ($(int.west)+(0,0.1)$) node [above, xshift=-2.3em, yshift=0.5em, align=center] {$\lambda_t \sim \pi(s_t)$};
    \draw[myarrow,draw=black!60] ($(int.west)+(0,-0.1)$) to [out=-150,in=-30] ($(policy.east)+(0,-0.1)$) node [below, xshift=2.3em, yshift=-0.5em, align=center] {$s_{t+1}$};
    
    %outer arrows
    \draw[myarrow,draw=black!60] ($(optimizer.east)+(0,0.1)$) to [in=150,out=30] ($(outer.west)+(0,0.1)$) node [above, xshift=-3.6em, yshift=0.6em, align=center] {update\\ $\pi, i$};
    \draw[myarrow,draw=black!60] ($(outer.west)+(0,-0.1)$) to [out=-150,in=-30] ($(optimizer.east)+(0,-0.1)$) node [below, xshift=3.6em, yshift=-0.7em, align=center] {Observations:\\ $(s_{t}, r_t, \lambda_t, s_{t+1})$};
    
    \draw[myarrow,draw=black!60] (space.south) -- (optimizer);
    \draw[myarrow,draw=black!60] (insts.north) -- (optimizer);
    
	\end{tikzpicture}
 	}
\caption{Interaction between optimizer, policy and all components of a DAC benchmark; latter in grey boxes.}
\label{fig:dacflow}%
\end{figure*}%

With DACBench, we strive for an easy-to-use, standardized and reproducible benchmark library that allows evaluating DAC on several, diverse benchmarks. 
To this end, we will first describe which components are needed to define a DAC benchmark, see Figure~\ref{fig:dacflow}, and then explain how we can make use of it to ensure our design objectives.

\subsection{Components of a DAC Benchmark}\label{components}

Inspired by the flexibility that the modelling as a cMDP allows
and the success of OpenAI's gym environments, 
each DACBench benchmark is modelled along these lines, with the following benchmark-specific design decisions.

\paragraph{Action Space $\mathcal{A}$} describes ways of modifying the current configuration. In the simplest case, the action space directly corresponds to the hyperparameter space, incl. all hyperparameter names and the corresponding ranges.

\paragraph{State Space $\mathcal{S}$} describes available information about the target algorithm state. This can be enriched by context information about the instance at hand.
We recommend that it is (i) cheap-to-compute information that is (ii) available at each step and (iii) measures the progress of the algorithm.

\paragraph{Target Algorithm with Transition Dynamics $\mathcal{T}_i$} implicitly defines which states $s_{t+1}$ are observed after hyperparameter configuration $\lambda_t$ is chosen in state $s_t$.
It is important to fix the target algorithm implementation (and all its dependencies) to ensure that this is reproducible.
An implicit design decision of a benchmark here is how long an algorithm should run before the next step description is returned. 

\paragraph{Reward Function $r_i$} provides a scalar signal of how well the algorithm can solve a given instance. 
It is an analogue to the cost function in AC and PIAC and should be the optimization target, e.g., prediction error, runtime or solution quality.

\paragraph{Instance Set $\mathcal{I}$} defines variants of a given problem that has to be solved s.t. the learned policy is able to generalize to new, but similar instances.\footnote{For simplicity, we only discuss the case of a set of training instances. In general, DACBench also supports instance generators s.t. the set of instances does not have to be fixed in advance.} To assess generalization performance, a training and test set of instances is required. In addition, instances can be described by instance features~\citep{bischl-aij16a} which facilitates learning of per-instance policies.

This fine granular view on benchmarks allows us on one hand to create a multitude of different benchmarks, potentially with different characteristics. On the other hand, a benchmark in DACBench is a specific instantiated combination of these components s.t. DACBench contributes to reproducible results.

\subsection{Practical Considerations \& Desiderata}\label{considerations}

DACBench provides a framework to implement the design decisions above with a focus on accessibility, reproducibility and supporting further research on DAC.

\textbf{Accessibility}
So far, applying a new DAC optimizer to a target problem domain requires domain knowledge to be able to interface with a potential algorithm.
Comparing optimizers across multiple benchmarks of varying characteristics often requires re-implementing or adapting parts of the optimizers to fit the different interfaces, hurting the consistency of the comparison and taking a lot of time and effort.

Similarly, developing and providing new and interesting benchmarks is challenging as, without a standardized interface, there is little guidance on how to do so.
Thus, domain experts wanting to provide a DAC benchmark of a target algorithm often construct their own interface, which can be time-consuming even with a background in MDPs.

Providing a standardized interface would alleviate the issues and facilitate moving DAC as a field forward.
Therefore, DACBench provides a common interface for benchmarks, based on OpenAI's gym API \citep{gym}, that makes interaction with DAC optimizers as simple as possible.
This interface is lightweight and intuitive to implement for experts from different domains, encouraging collaboration in the creation of new benchmarks and optimizers.
It also allows domain experts to modify existing benchmarks with little effort and minimal knowledge of the base code to create new and interesting variations of known benchmarks, see Appendix~\ref{sec:mod}.

\textbf{Reproducibility} 
As discussed before, adapting an algorithm for DAC can be challenging as there are many design decisions involved. 
On one hand, to allow studies of new DAC characteristics, we believe it is important to give researchers the flexibility to adjust these components.
Therefore, we do not want to propose a framework that fixes too many decision points as it could restrict important future research.
On the other hand, we believe there is a need for standardized benchmarks to facilitate comparing different methods as well as reproducing research results.
For this purpose, all design decisions of the original experiments should be reproducible.
To this end, DACBench includes a mechanism to customize as many of these design decisions as possible, but also to record them such that other researchers can reproduce the experiments (for more details, see Appendix~\ref{app:implementation}).

\textbf{Facilitating Further Research} 
Lastly, DACBench supports researchers by providing resources needed to work on DAC problems as well as thorough documentation of the design decisions of each benchmark.
As existing benchmarks are often not documented very well, working with them requires thorough study of the code base.
Instead, DACBench provides all important details about individual benchmarks in a concise manner through comprehensive documentation.

Furthermore, DACBench provides quality of life components like structured logging and visualization that make working with DACBench seamless.
The logging system gives users the option to save a variety of details like the policies or state information for later analysis.
Further, the built-in visualization tools make evaluating experiments easy (examples include Figures~\ref{fig:modea_seeds}, \ref{fig:fd_seeds} and \ref{fig:plots}) and can directly use the data provided by the logging system.

These considerations contribute to driving open research on DAC forward by ensuring easy reproducibility of experiments, usability for a diverse audience and sharing of experiment configurations.
By adopting a simple yet modular interface, we improve general accessibility to the field as well as the ability to continuously evolve DAC benchmarks.

\subsection{Six Initial Diverse DAC Benchmarks}
We propose six initial benchmarks for DACBench from different domains and with different challenges (for in-depth descriptions, see Appendix \ref{app:benchmarks}).
We believe these present a versatile set of problems both for testing DAC methods across diverse benchmarks and developing new approaches. 

\textbf{Sigmoid \& Luby} 
\citep{biedenkapp-ecai20} are time series approximation tasks with no underlying target algorithm. 
These artificial benchmarks run very quickly, their optimal policies can be computed efficiently for all possible instances (i.e. transformations of the functions themselves) and it is easy to generate instance sets for a wide range of difficulties. 
Therefore, Sigmoid and Luby are ideal for DAC developers, e.g. to verify that agents can learn the optimal policy or slowly ramp up the instance heterogeneity in order to test its generalization capabilities. 

\textbf{FastDownward} 
\citep{helmert-jair06a} is a state-of-the-art AI Planner, which gives rise to a more complex benchmark. The task here is to select the search heuristic at each step on a specific problem family with two complementary heuristics.
This can be considered one of the easier benchmarks even though significant performance gains on competition domains are possible with four commonly used heuristics \citep{speck-icaps21}. 
The basic instance set we provide includes optimal policy information as an upper performance bound.

\textbf{CMA-ES} 
\citep{hansen-ec03} is an evolutionary strategy, where the DAC task is to adapt the algorithm's steps size~\citep{shala-ppsn20} when solving BBOB functions. However, finding a good solution in this continuous space is potentially harder than the discrete heuristic selection in FastDownward. While optimal policies are unknown for this benchmark, there is a strong established dynamic baseline in CSA~\citep{chotard-ppsn12}.

\begin{figure}[tbp]
    \centering
    \subfloat[]{
    \includegraphics[scale=0.3]{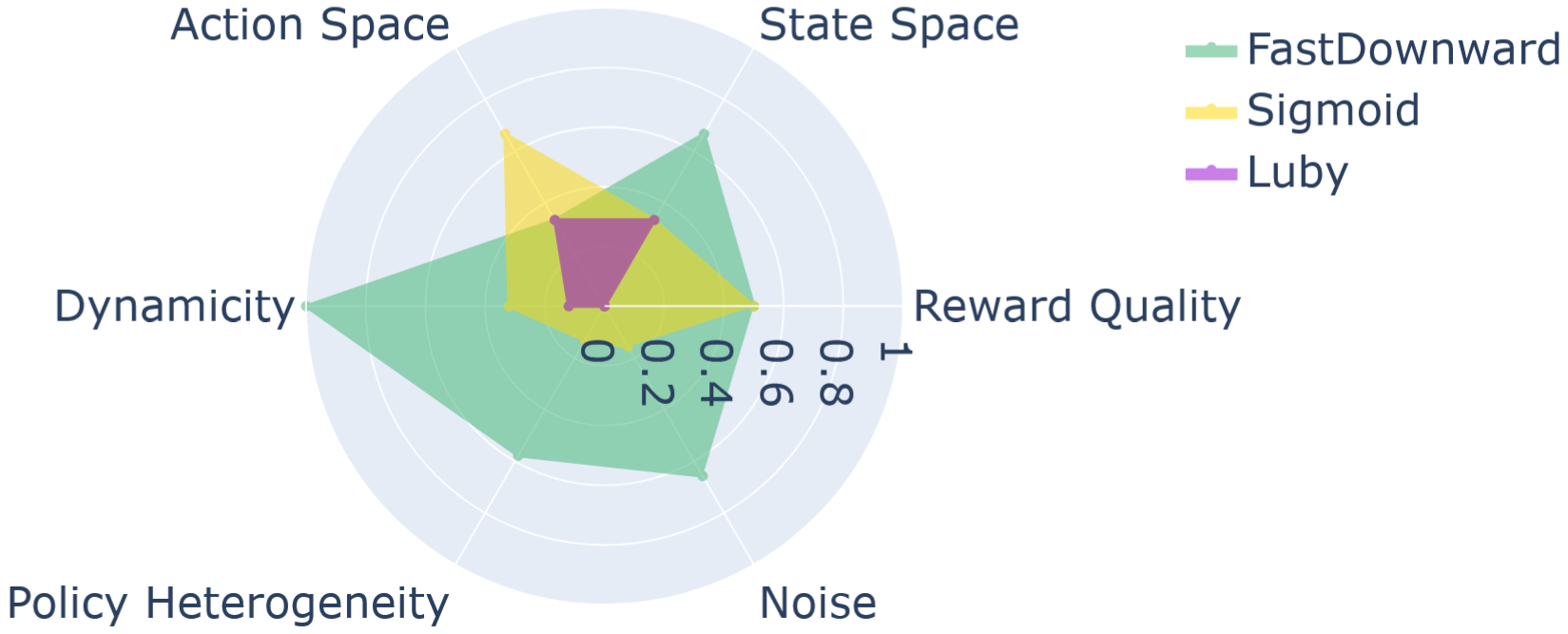}}
    
    \subfloat[]{
    \includegraphics[scale=0.3]{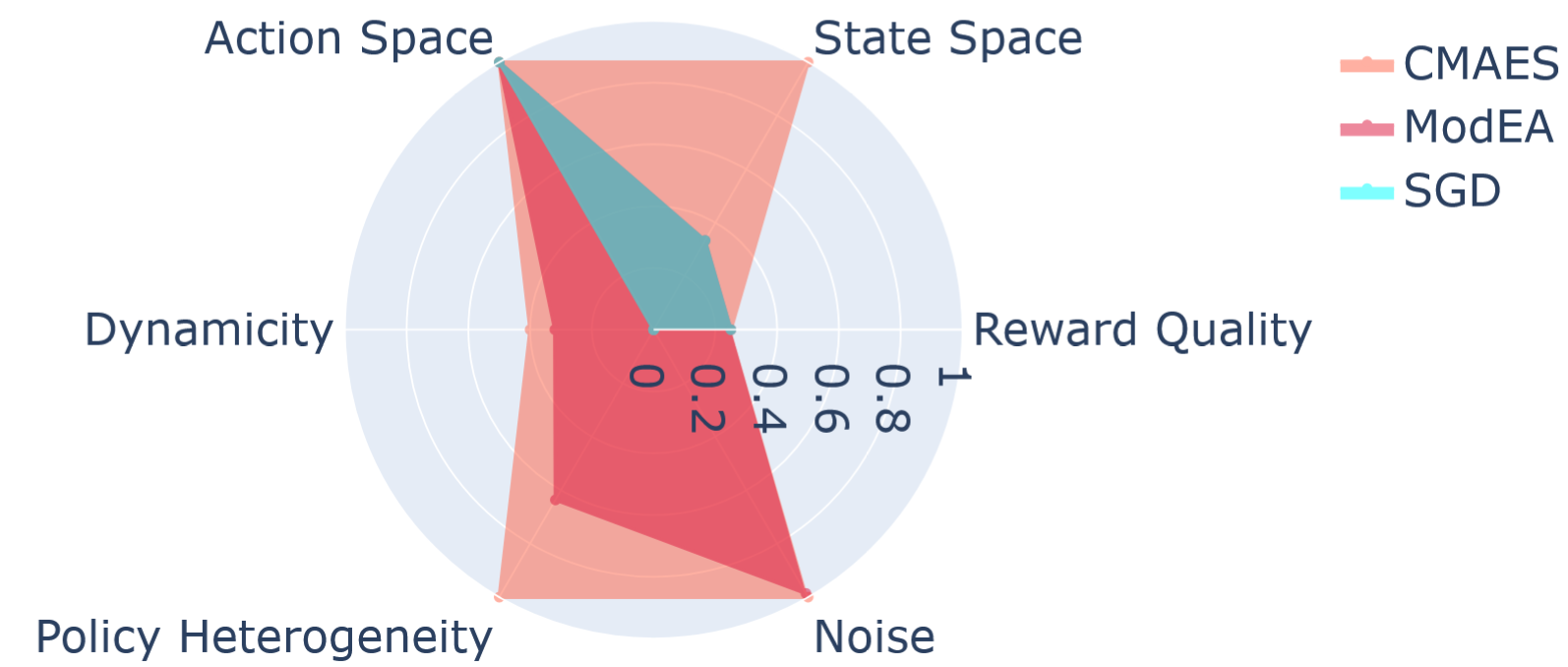}}
    \caption{Ranked comparison of difficulty dimensions in DACBench benchmarks. Lower values correspond to easier characteristics.}
    \label{fig:spider}
\end{figure}

\textbf{ModEA} 
includes an example of dynamic algorithm selection for variants of CMA-ES on BBOB functions~\citep{vermetten-gecco19}. In contrast to the CMA-ES benchmark, a combination of $11$ EA elements with two to three options each are chosen in each step; this combination makes up the final algorithm. This multi-dimensional, large action space makes the problem very complex. So we expect this to be a hard benchmark, possibly too hard for current DAC approaches to efficiently determine an effective DAC policy.

\textbf{SGD-DL} 
adapts the learning rate of a small neural network learning a simple image classification task ~\citep{daniel-aaai16}. 
The small network size allows for efficient development and benchmarking of new DAC approaches. By varying the instance (dataset-seed pairs) and the network architecture, this benchmark nevertheless opens up ample possibility to grow ever harder as DAC advances.

\section{Empirical Insights Gained from DACBench}
In order to study our benchmarks, we discuss dimensions of difficulty which are relevant to the DAC setting.
To provide insights into how our benchmarks behave in these dimensions, we use static policies, known dynamic baselines and random dynamic policies to explore their unique challenges. 

\subsection{Setup}
To show how our benchmarks behave in practice, we mainly use the static and random policies built into DACBench and, where possible, make use of optimal policies. 
All of them were run for $10$ seeds with at most $1\,000$ steps on each instance. 
For benchmarks with a discrete action space, static policies cover all the actions.
The two benchmarks with continuous action spaces, CMA-ES and SGD-DL were run with $50$ static actions each, distributed uniformly over the action space.
For details on the hardware used, refer to Appendix~\ref{app:hardware}.

\subsection{Coverage of Difficulty Dimensions}
Similar to \citet{biedenkapp-ecai20}, we identified six core challenges of learning dynamic configuration policies to characterize our benchmarks. 
For comparison's sake, we define a scale for each attribute and measure these on our benchmarks.
These dimensions of difficulty are:
(i) \emph{State} and (ii) \emph{action space size} increase the difficulty of the problem by varying information content, requiring the agent to learn what state information is relevant and which regions in the action space are useful.
(iii) \emph{Policy heterogenity} quantifies how successful different policies are across all instances.
(iv) \emph{Reward quality} refers to the information content of the given reward signal.
(v) \emph{Noise} can disturb the training process through noisy transitions or rewards.
Lastly, (vi) \emph{dynamicity} shows how frequently the action should be changed, i.e. how complex well-performing policies need to be. See Appendix~\ref{app:dimension} for details.

Figure~\ref{fig:spider} shows how the benchmarks compare with respect to these dimensions of difficulty. 
While the reward quality is not fully covered, we cover all other dimensions well, with at least a very, moderately and not especially difficult benchmark in each.
Additionally, all DACBench benchmarks show a different profile.
The data shows that Luby could be considered the easiest of the six, with little noise or policy heterogeneity and a relatively low dynamicity score, requiring only infrequent action changes.
SGD-DL's footprint looks similar, though its continuous action space makes for a difficulty spike in that category.
While Sigmoid's reward function obscures quite a bit of information, it is not very difficult in the other dimensions.
FastDownward on the other hand leads the dynamicity dimension by far, showing a need for more active control.
It is also fairly challenging with regard to noise and policy heterogeneity.
CMA-ES is even more difficult in these, while also having the largest state space.
A more informative reward and lower dynamicity contrast it and other benchmarks.
ModEA's difficulty, on the other hand, seems similar except for the challenge of a continuous state space. 

While this shows that our benchmark set covers all of our dimensions of difficulty with the exception of reward quality fairly well, we will continue to explore the dimensions of noise, policy heterogeneity and dynamicity in greater detail in order to give a more detailed impression of how these dimensions are expressed.

\begin{figure}
    \centering
        \includegraphics[width=0.45\textwidth]{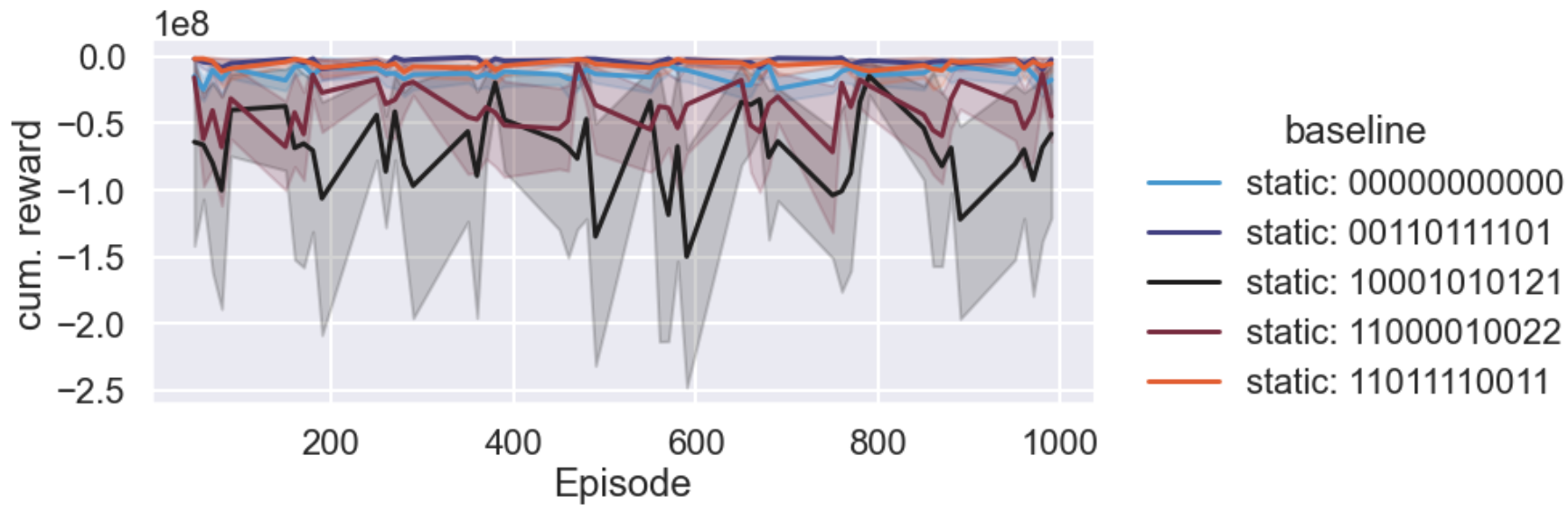}
    \caption{Performance of $5$ static ModEA policies with $95\%$ confidence interval. The legend shows which components of ModEA were used.}
    \label{fig:modea_seeds}
\end{figure}

\begin{figure}
    \centering
    \includegraphics[width=0.4\textwidth]{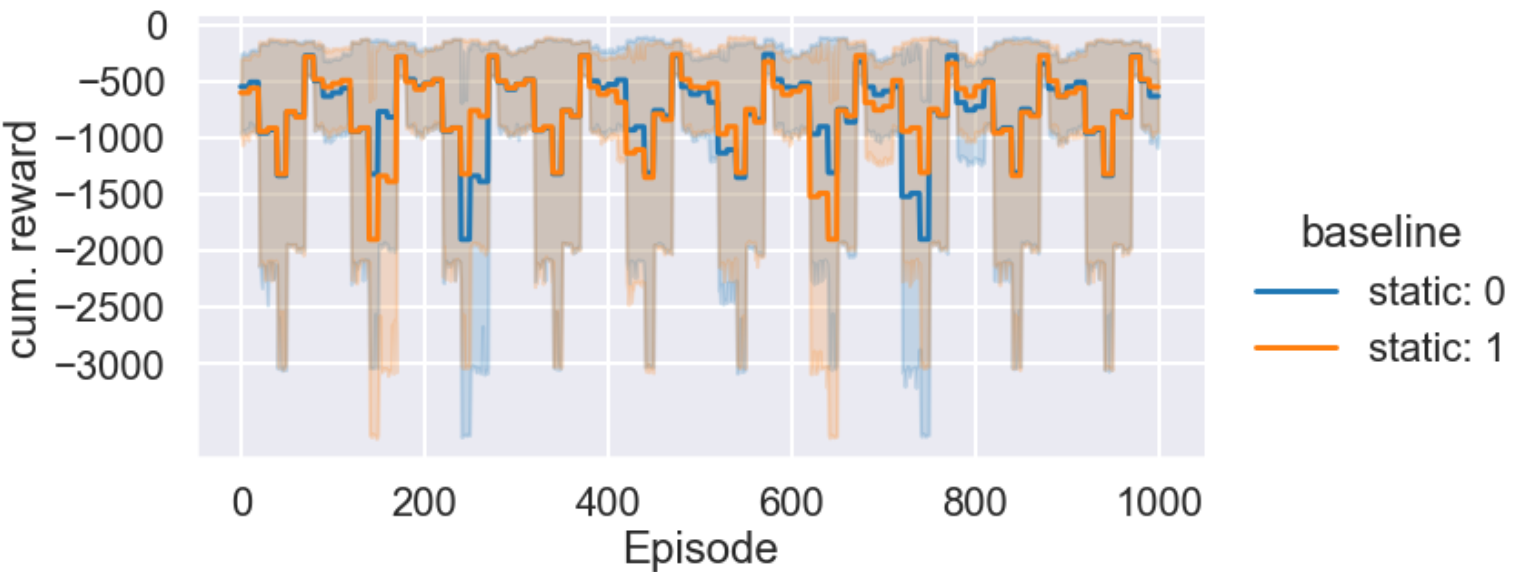}
    \caption{Comparison of average performance of static FastDownward policies with $95\%$ confidence interval.}
    \label{fig:fd_seeds}
\end{figure}

\subsection{Degree of Randomness}
To show how randomness is expressed in our benchmarks, we investigate its effects on FastDownward and ModEA.

We quantified randomness by using the standard deviation of the cumulative reward between different seeds for the same actions, each repeated $10$ times.
ModEA was one of the benchmarks that had a very high relative standard deviation and thus a very high noise score, see Figure~\ref{fig:modea_seeds}.
While static policies from different parts of the action space vary in performance, their confidence intervals grow much larger the worse they perform.
This is to be expected, as policies with a high reward have found EA components that quickly find the optimal solution of the black-box function at hand.
If the resulting EA cannot find a solution quickly, the individuals in each generation will have very different proposed solutions, thus resulting in unstable performance.
So even though ModEA contains quite a bit of noise, the noise is heteroscedastic, i.e., it is not evenly distributed across the policy space, providing an additional challenge.

FastDownward, on the other hand, also has a high rating in the noise category, but the way its noise is distributed is quite different, see Figure~\ref{fig:fd_seeds}.
W.r.t. the average performance of both static policies, the $95\%$ confidence interval is up to twice as large as the performance value itself.
In contrast to ModEA, the noise is large but likely homoscedastic.

\begin{figure}[tbp]
    \centering
    %\subfloat[]{\label{fig:cma_a} %ML: requires unnecessary space
    \includegraphics[scale=0.21]{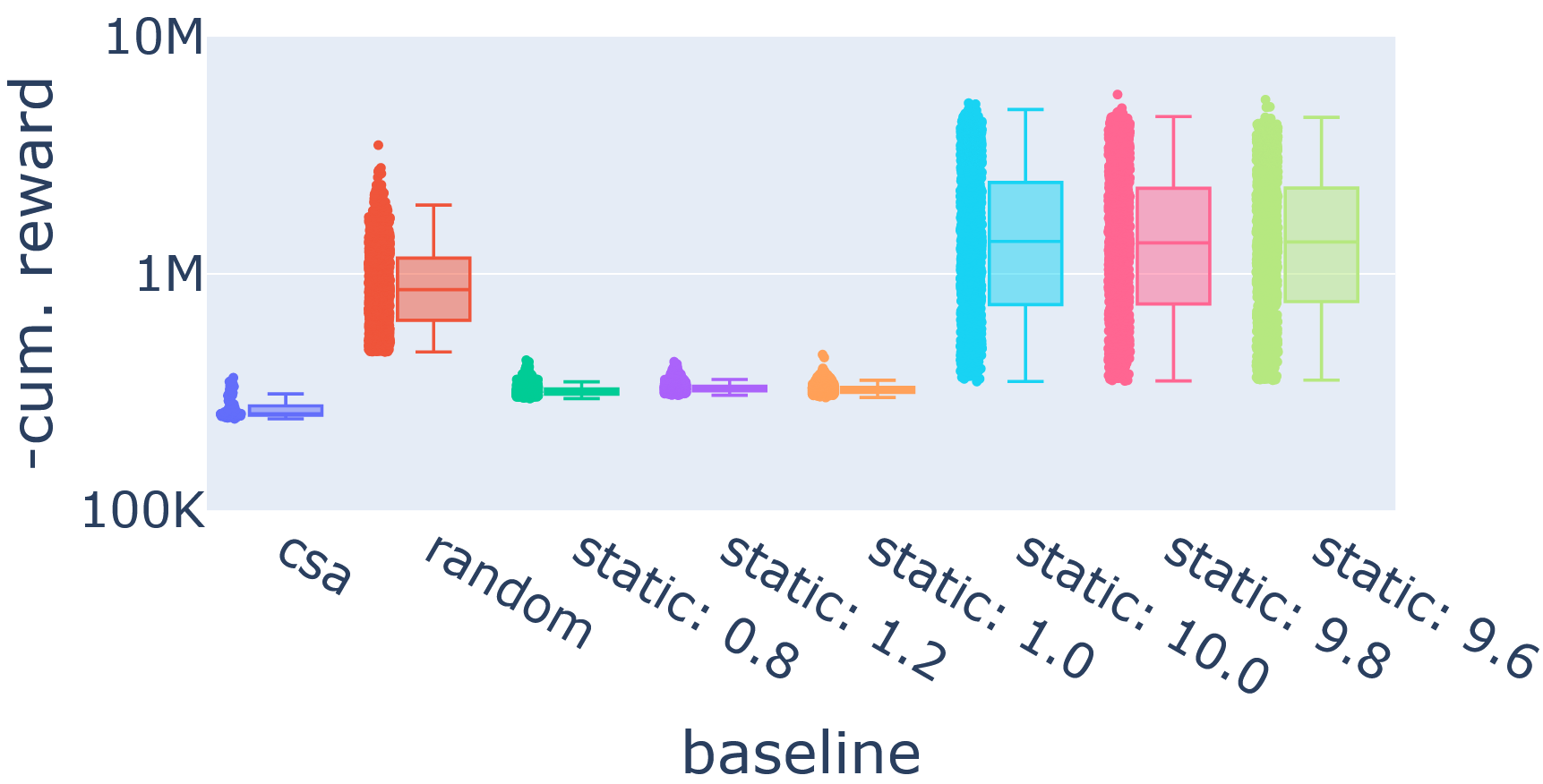}%}
    %\subfloat[]{
    %\label{fig:cma_b}
    \includegraphics[scale=0.21]{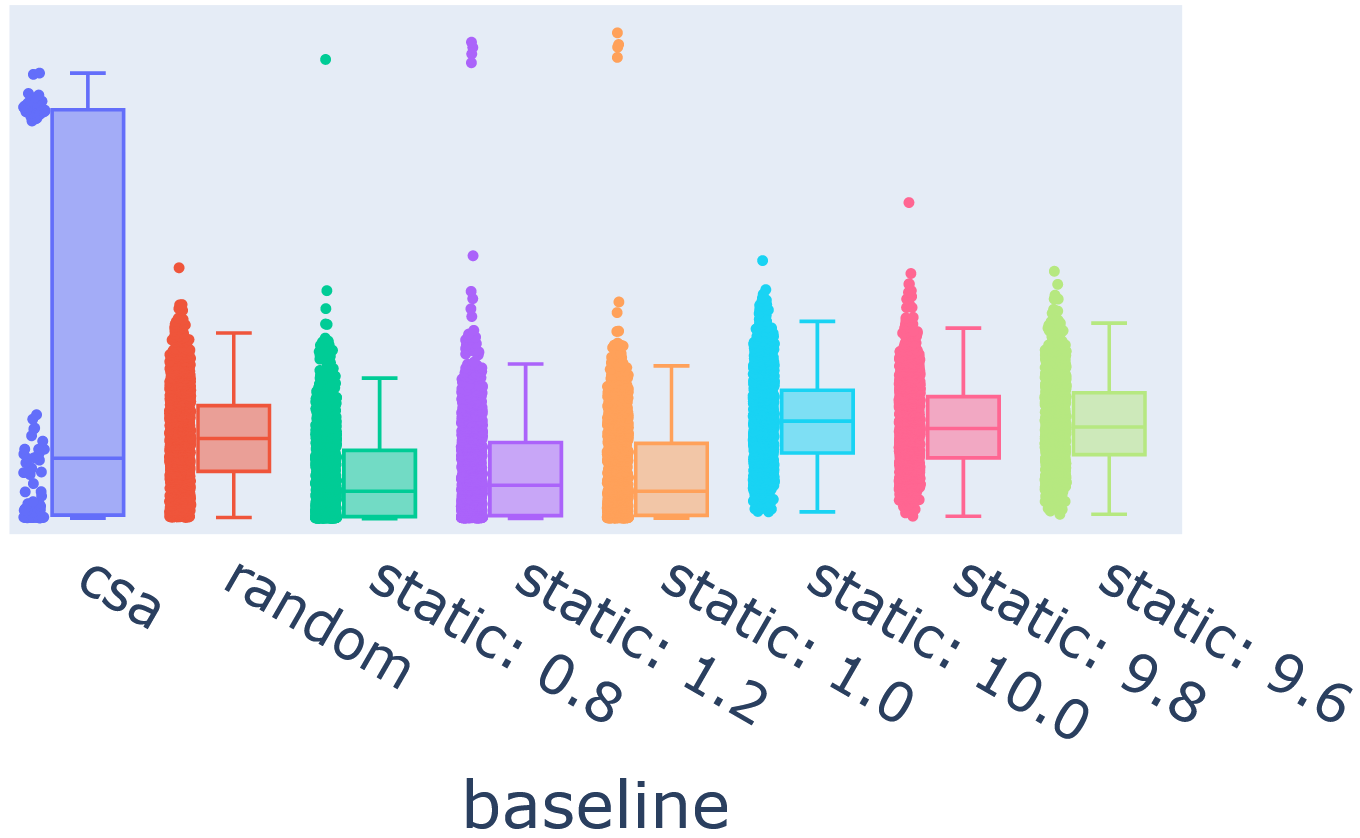}%}
    \caption{Policy evaluation of CMA benchmark on Schaffers (left) and Ellipsoid (right) functions (with 3 best and worst static policies).}
    \label{fig:cma_functions}
\end{figure}

\begin{figure}[tbp]
    \centering
    \includegraphics[scale=0.3]{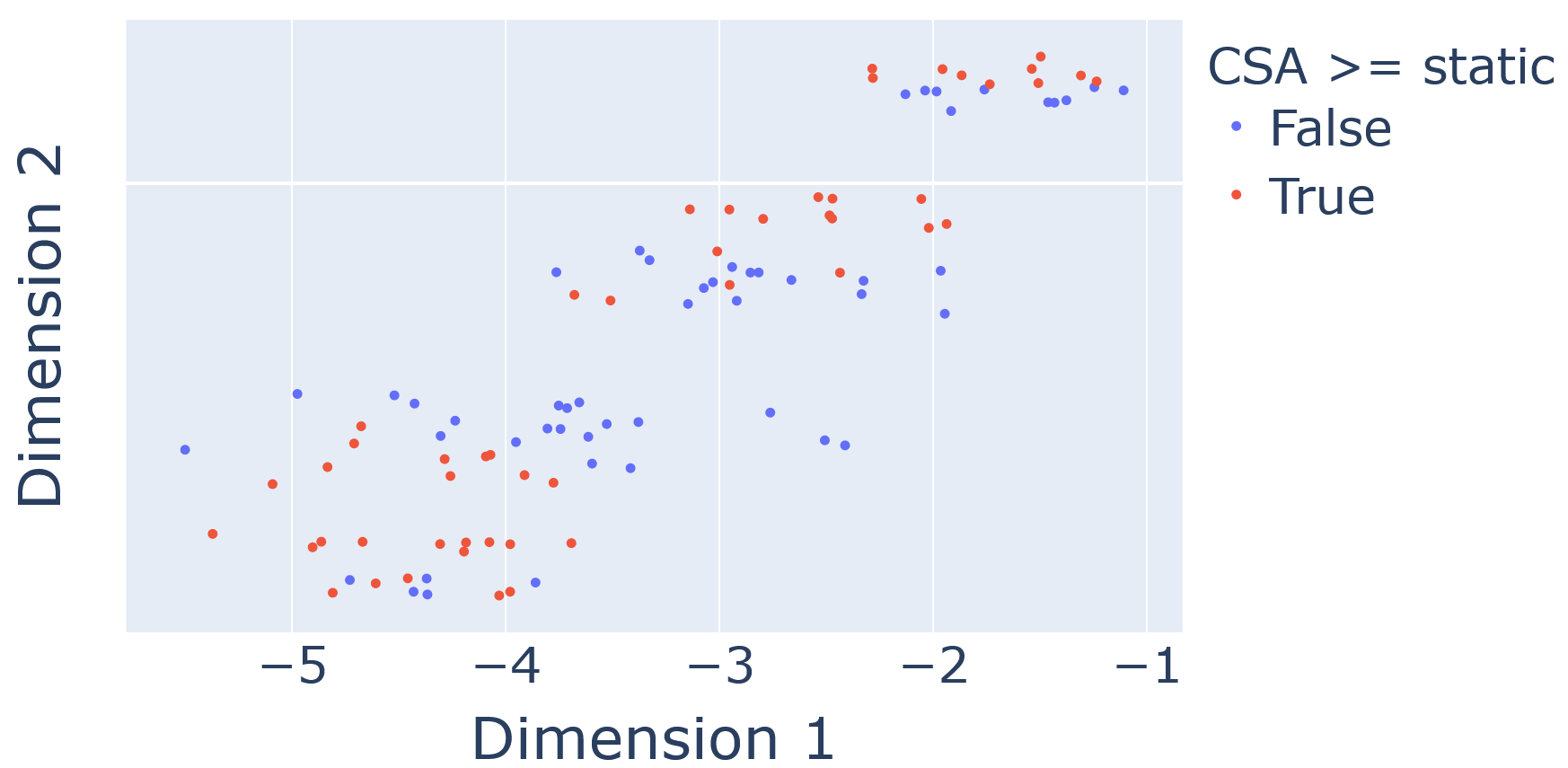}
    \caption{Algorithm footprint t-SNE plot of CMA-ES instances showing where CSA outperforms all static policies.}
    \label{fig:cma_tsne}
\end{figure}

%\begin{figure}[h!]
\begin{figure*}[h!]
    \centering
    \includegraphics[width=0.31\textwidth]{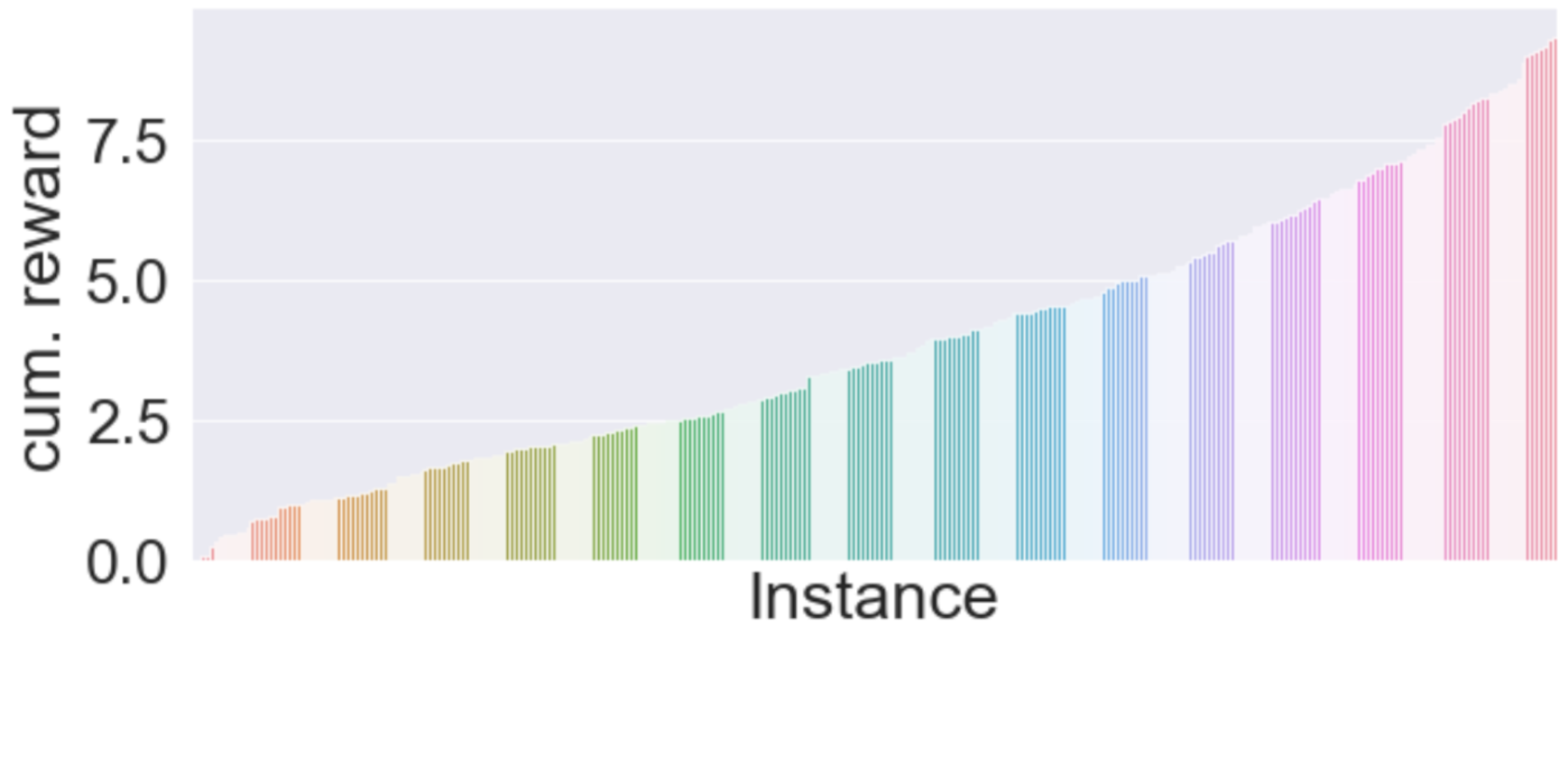} \includegraphics[width=0.32\textwidth]{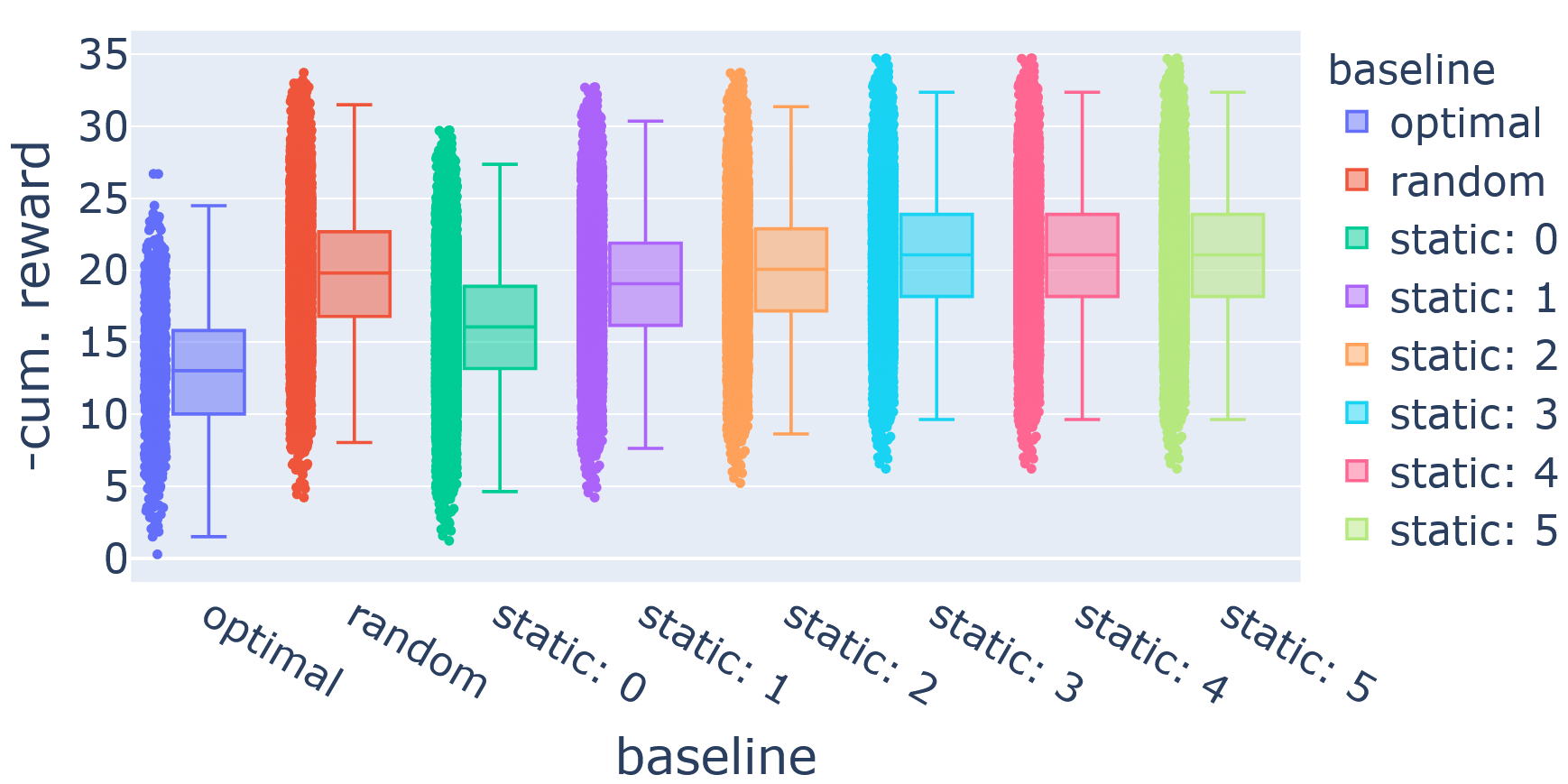}
    \includegraphics[width=0.32\textwidth]{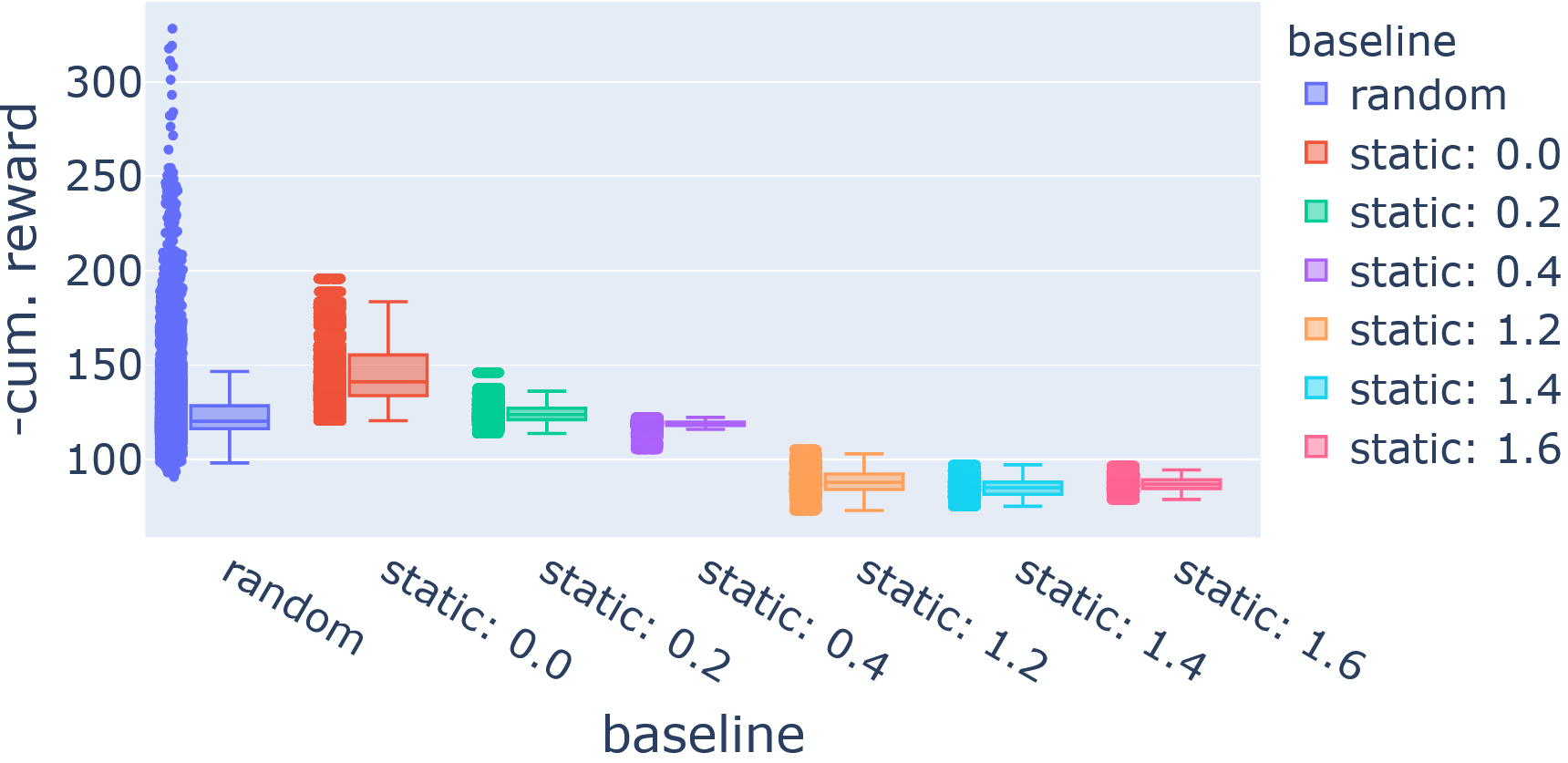}
    \caption{\textbf{Left:} Best possible reward for each sigmoid instance. \textbf{Middle:} Static and dynamic policies on Luby. The reward is $0$ if the agent guesses the correct sequence element, $-1$ otherwise. \textbf{Right:} Static (best and worst 3) and dynamic policies on SGD-DL with $\lambda = 10^{-x}$. The reward here is the validation loss (negative log-likelihood).}
    \label{fig:plots}
\end{figure*}
%\end{figure}
\subsection{Effect of Instances}
To investigate the effect instances have on our benchmarks, we examine CMA-ES, which showed the highest policy heterogeneity above, and Sigmoid, for which we can compute the optimal policy.
CMA-ES and ModEA both operate on instance sets comprised of different function classes between which we can clearly see very different behaviour.
The Schaffers function (see Figure \ref{fig:cma_functions} left) illustrates that the hand-designed CSA is indeed a good dynamic policy; it outperforms all other static and random policies.

In contrast, CSA performs much worse on the Ellipsoid function (Figure \ref{fig:cma_functions} right).
Using the probability estimation proposed by \cite{shala-ppsn20} based on the Wilcoxon rank sum test, CSA's probability of outperforming any given static policy is $74.6\%$ overall; also shown on a per-instance level in the algorithm footprint~\citep{smith-miles-cor14} in Figure~\ref{fig:cma_tsne}.
While this shows that CSA's dynamic control policy is preferred on most of CMA-ES instance space, there are also regions that require a different approach, underlining the importance of instance dependent methods.

On the Sigmoid benchmark we see that performance differences between instances persist even for the optimal policy (see Figure~\ref{fig:plots} left). While it performs very well on some instances, this is far from the case for all of them. 
Indeed, while it is possible to gain the best possible reward of $10$ on some instances, there is an almost even distribution of rewards between the maximum and minimum cumulative reward.

Overall, different instances can have a significant influence on the overall performance, both in terms of which policies are successful on them and how well an agent can do.

\subsection{Is Dynamic Better than Static?}
Even though we have empirical evidence of DAC agents surpassing static baselines for all of our benchmarks \citep{daniel-aaai16,vermetten-gecco19,biedenkapp-ecai20,shala-ppsn20,speck-icaps21}, we analyse and compare the performance of dynamic and static policies on our benchmarks.
This way we can estimate the difficulty both in finding a good dynamic policy that surpasses a simple random one but also the difficulty of outperforming the static policies.
Insights into the relationship between static and dynamic policies can highlight characteristics of a benchmark, give upper and lower performance bounds and show the effect size we can expect from DAC approaches in the future.

Our evaluation clearly shows that the benchmarks have a very different sensitivity to dynamic policies. 
In Luby (Figure~\ref{fig:plots} middle) we can see that the most common elements of the Luby sequence, elements one and two, outperform the dynamic random policy. 
As $50\%$ of the Luby sequence consist of the first element and $25\%$ of the second, this is the expected behaviour.
Therefore it also makes sense that the optimal policy outperforms all other policies.
The random policy does not perform very well, showing that there is a lot of room to improve over it and subsequently over the static policies.

Similarly, the random policy of SGD-DL outperforms some of the worst static policies on average, but does very poorly compared to them on many occasions (see Figure~\ref{fig:plots} right).
Improving over the best static policies here will therefore be much harder for a DAC agent.
This is also an example of the fact that dynamically adapting hyperparameters can outperform static settings, as \cite{daniel-aaai16} showed for this setting, but the region of well-performing dynamic policies seem to be much smaller than for Luby above.
This is the reason for the benchmark's low dynamicity rating. Unlike e.g. FastDownward, which favors frequent action changes regardless of their quality, SGD-DL requires a more subtle approach with more consistency and carefully selected actions.

Therefore, we believe dynamicity will play a large role in how DAC methods should approach benchmarks. 
While choosing a new action each step for SGD-DL can of course be learned successfully over time, it is a much harder task than Luby.
Methods keeping actions for a number of steps at a time may have better success here~\citep{vermetten-gecco19}.

\section{Conclusion}
We propose DACBench, a standardized benchmark suite for dynamic algorithm configuration (DAC). 
With it, we provide a framework to configure DAC benchmarks that both enables reproducibility and easy modifications, ensuring that DACBench can help evolve DAC benchmarks further.
For example, we plan to extend the FastDownward benchmark beyond single domains and include existing instance features from e.g. Exploratory Landscape Analysis (ELA) for CMA-ES and ModEA.
Furthermore, DACBench is easily extendable and we will add new benchmarks, developed by us and the community.
As an incentive for researchers to tackle some of the most important difficulties in solving DAC, we provide challenges for several dimensions of hardness.
In order to assist in developing these new approaches, we also include tools for tracking important metrics and visualization, making DACBench very easy to use without knowledge of the target domains.
Overall, we believe DACBench will make DAC more accessible to interested researchers, make existing DAC approaches more easily comparable and provide a direction for research into new methods. For future work, we plan to build surrogate benchmarks, similar to \cite{eggensperger-mlj18a} for AC and \cite{siems-arxiv20a} for NAS, to enable DAC benchmarking with minimal computational overhead and minimized CO$_2$ footprint.

\section*{Acknowledgements}
We thank Gresa Shala, David Speck and Rishan Senanayake for their contributions to the CMA-ES, FastDownward and SGD-DL benchmarks respectively.
Theresa Eimer and Marius Lindauer acknowledge funding by the German Research Foundation (DFG) under LI 2801/4-1.
All authors acknowledge funding by the Robert Bosch GmbH.

{\small
\bibliographystyle{named}
\bibliography{bib/shortstrings,bib/lib,bib/dacbench,bib/shortproc}}

\begin{thebibliography}{}

\bibitem[\protect\citeauthoryear{Almutairi \bgroup \em et al.\egroup
  }{2016}]{almutairi-iscis16}
A.~Almutairi, E.~{\"{O}}zcan, A.~Kheiri, and W.~Jackson.
\newblock Performance of selection hyper-heuristics on the extended hyflex
  domains.
\newblock In {\em Proc. of {ISCIS}}, pages 154--162, 2016.

\bibitem[\protect\citeauthoryear{Andrychowicz \bgroup \em et al.\egroup
  }{2016}]{andrychowicz-neurips16}
M.~Andrychowicz, M.~Denil, S.~G. Colmenarejo, M.~W. Hoffman, D.~Pfau,
  T.~Schaul, and N.~de~Freitas.
\newblock Learning to learn by gradient descent by gradient descent.
\newblock In {\em Proc. of {N}eur{IPS}}, pages 3981--3989, 2016.

\bibitem[\protect\citeauthoryear{Ans{\'o}tegui \bgroup \em et al.\egroup
  }{2009}]{ansotegui-cp09a}
C.~Ans{\'o}tegui, M.~Sellmann, and K.~Tierney.
\newblock A gender-based genetic algorithm for the automatic configuration of
  algorithms.
\newblock In {\em Proc. of {CP}'09}, pages 142--157, 2009.

\bibitem[\protect\citeauthoryear{Ans{\'{o}}tegui \bgroup \em et al.\egroup
  }{2016}]{ansotegui-aij16}
C.~Ans{\'{o}}tegui, J.~Gab{\`{a}}s, Y.~Malitsky, and M.~Sellmann.
\newblock Maxsat by improved instance-specific algorithm configuration.
\newblock {\em AIJ}, 235:26--39, 2016.

\bibitem[\protect\citeauthoryear{Biedenkapp \bgroup \em et al.\egroup
  }{2020}]{biedenkapp-ecai20}
A.~Biedenkapp, H.~F. Bozkurt, T.~Eimer, F.~Hutter, and M.~Lindauer.
\newblock Dynamic {A}lgorithm {C}onfiguration: {F}oundation of a {N}ew
  {M}eta-{A}lgorithmic {F}ramework.
\newblock In {\em Proc. of {ECAI}}, pages 427--434, 2020.

\bibitem[\protect\citeauthoryear{Bischl \bgroup \em et al.\egroup
  }{2016}]{bischl-aij16a}
B.~Bischl, P.~Kerschke, L.~Kotthoff, M.~Lindauer, Y.~Malitsky,
  A.~Frech\'{e}tte, H.~Hoos, F.~Hutter, K.~Leyton-Brown, K.~Tierney, and
  J.~Vanschoren.
\newblock {ASlib}: A benchmark library for algorithm selection.
\newblock {\em AIJ}, pages 41--58, 2016.

\bibitem[\protect\citeauthoryear{Brockman \bgroup \em et al.\egroup
  }{2016}]{gym}
G.~Brockman, V.~Cheung, L.~Pettersson, J.~Schneider, J.~Schulman, J.~Tang, and
  W.~Zaremba.
\newblock {OpenAI Gym}.
\newblock {\em CoRR}, abs/1606.01540, 2016.

\bibitem[\protect\citeauthoryear{Chen \bgroup \em et al.\egroup
  }{2017}]{chen-icml17}
Y.~Chen, M.~Hoffman, S.~Colmenarejo, M.~Denil, T.~Lillicrap, M.~Botvinick, and
  N.~de~Freitas.
\newblock Learning to learn without gradient descent by gradient descent.
\newblock In {\em Proc. of {ICML}}, pages 748--756, 2017.

\bibitem[\protect\citeauthoryear{Chotard \bgroup \em et al.\egroup
  }{2012}]{chotard-ppsn12}
A.~Chotard, A.~Auger, and N.~Hansen.
\newblock Cumulative step-size adaptation on linear functions.
\newblock In {\em Proc. of {PPSN}}, 2012.

\bibitem[\protect\citeauthoryear{Daniel \bgroup \em et al.\egroup
  }{2016}]{daniel-aaai16}
C.~Daniel, J.~Taylor, and S.~Nowozin.
\newblock Learning step size controllers for robust neural network training.
\newblock In {\em Proc. of {AAAI}}, pages 1519--1525, 2016.

\bibitem[\protect\citeauthoryear{Doerr and Doerr}{2018}]{doerr-algo18}
B.~Doerr and C.~Doerr.
\newblock Optimal static and self-adjusting parameter choices for the
  (1+({\(\lambda\)}, {\(\lambda\)})) genetic algorithm.
\newblock {\em Algorithmica}, 80(5):1658--1709, 2018.

\bibitem[\protect\citeauthoryear{Doerr \bgroup \em et al.\egroup
  }{2018}]{IOHprofiler}
C.~Doerr, H.~Wang, F.~Ye, S.~van Rijn, and T.~B{\"a}ck.
\newblock Iohprofiler: A benchmarking and profiling tool for iterative
  optimization heuristics.
\newblock {\em arXiv e-prints:1810.05281}, 2018.

\bibitem[\protect\citeauthoryear{Eggensperger \bgroup \em et al.\egroup
  }{2013}]{eggensperger-bayesopt13}
K.~Eggensperger, M.~Feurer, F.~Hutter, J.~Bergstra, J.~Snoek, H.~Hoos, and
  K.~Leyton-Brown.
\newblock Towards an empirical foundation for assessing {Bayesian} optimization
  of hyperparameters.
\newblock In {\em {NeurIPS} Workshop on {B}ayesian Optimization in Theory and
  Practice (BayesOpt'13)}, 2013.

\bibitem[\protect\citeauthoryear{Eggensperger \bgroup \em et al.\egroup
  }{2018}]{eggensperger-mlj18a}
K.~Eggensperger, M.~Lindauer, H.~H. Hoos, F.~Hutter, and K.~Leyton{-}Brown.
\newblock Efficient benchmarking of algorithm configurators via model-based
  surrogates.
\newblock {\em Machine Learning}, 107(1):15--41, 2018.

\bibitem[\protect\citeauthoryear{Gupta \bgroup \em et al.\egroup
  }{2018}]{gupta-neurips18}
A.~Gupta, R.~Mendonca, Y.~Liu, P.~Abbeel, and S.~Levine.
\newblock Meta-reinforcement learning of structured exploration strategies.
\newblock In {\em Proc. of {NeurIPS}}, pages 5307--5316, 2018.

\bibitem[\protect\citeauthoryear{Hallak \bgroup \em et al.\egroup
  }{2015}]{hallak-corr15}
A.~Hallak, D.~Di Castro, and S.~Mannor.
\newblock Contextual markov decision processes.
\newblock {\em CoRR}, abs/1502.02259, 2015.

\bibitem[\protect\citeauthoryear{Hansen \bgroup \em et al.\egroup
  }{2003}]{hansen-ec03}
Nikolaus Hansen, Sibylle~D. M{\"{u}}ller, and Petros Koumoutsakos.
\newblock Reducing the time complexity of the derandomized evolution strategy
  with covariance matrix adaptation {(CMA-ES)}.
\newblock {\em Evolutionary Computing}, 11(1):1--18, 2003.

\bibitem[\protect\citeauthoryear{Hansen \bgroup \em et al.\egroup
  }{2020}]{hansen-oms2020}
N.~Hansen, A.~Auger, R.~Ros, O.~Mersmann, T.~Tu{\v s}ar, and D.~Brockhoff.
\newblock {COCO}: A platform for comparing continuous optimizers in a black-box
  setting.
\newblock {\em Optimization Methods and Software}, 2020.

\bibitem[\protect\citeauthoryear{Helmert}{2006}]{helmert-jair06a}
M.~Helmert.
\newblock The fast downward planning system.
\newblock {\em JAIR}, 26:191--246, 2006.

\bibitem[\protect\citeauthoryear{Houthooft \bgroup \em et al.\egroup
  }{2018}]{houthooft-neurips18}
R.~Houthooft, Y.~Chen, P.~Isola, B.~Stadie, F.~Wolski, J.~Ho, and Pieter
  Abbeel.
\newblock Evolved policy gradients.
\newblock In {\em Proc. of {NeurIPS}}, pages 5405--5414, 2018.

\bibitem[\protect\citeauthoryear{Hutter \bgroup \em et al.\egroup
  }{2009}]{hutter-jair09a}
F.~Hutter, H.~Hoos, K.~Leyton-Brown, and T.~St{\"u}tzle.
\newblock Param{ILS}: An automatic algorithm configuration framework.
\newblock {\em JAIR}, 36:267--306, 2009.

\bibitem[\protect\citeauthoryear{Hutter \bgroup \em et al.\egroup
  }{2011}]{hutter-lion11a}
F.~Hutter, H.~Hoos, and K.~Leyton-Brown.
\newblock Sequential model-based optimization for general algorithm
  configuration.
\newblock In {\em Proc. of {LION}}, pages 507--523, 2011.

\bibitem[\protect\citeauthoryear{Hutter \bgroup \em et al.\egroup
  }{2014}]{hutter-lion14a}
F.~Hutter, M.~L\'{o}pez-Ib\'{a}nez, C.~Fawcett, M.~Lindauer, H.~Hoos,
  K.~Leyton-Brown, and T.~St\"utzle.
\newblock {AClib}: a benchmark library for algorithm configuration.
\newblock In {\em Proc. of {LION}}, pages 36--40, 2014.

\bibitem[\protect\citeauthoryear{Hutter \bgroup \em et al.\egroup
  }{2017}]{hutter-aij17a}
F.~Hutter, M.~Lindauer, A.~Balint, S.~Bayless, H.~Hoos, and K.~Leyton-Brown.
\newblock The configurable {SAT} solver challenge ({CSSC}).
\newblock {\em AIJ}, 243:1--25, 2017.

\bibitem[\protect\citeauthoryear{Hutter \bgroup \em et al.\egroup
  }{2019}]{hutter-book19a}
F.~Hutter, L.~Kotthoff, and J.~Vanschoren, editors.
\newblock {\em Automated Machine Learning: Methods, Systems, Challenges}.
\newblock Springer, 2019.
\newblock Available for free at http://automl.org/book.

\bibitem[\protect\citeauthoryear{L{\'{o}}pez{-}Ib{\'{a}}{\~{n}}ez \bgroup \em
  et al.\egroup }{2016}]{lopez-ibanez-orp16}
M.~L{\'{o}}pez{-}Ib{\'{a}}{\~{n}}ez, J.~Dubois-Lacoste, L.~Perez Caceres,
  M.~Birattari, and T.~St{\"{u}}tzle.
\newblock The irace package: Iterated racing for automatic algorithm
  configuration.
\newblock {\em Operations Research Perspectives}, 3:43--58, 2016.

\bibitem[\protect\citeauthoryear{Loshchilov and
  Hutter}{2017}]{loshchilov-iclr17a}
I.~Loshchilov and F.~Hutter.
\newblock Sgdr: Stochastic gradient descent with warm restarts.
\newblock In {\em Proc. of {ICLR}}, 2017.

\bibitem[\protect\citeauthoryear{Ochoa \bgroup \em et al.\egroup
  }{2012}]{burke-mista09}
G.~Ochoa, M.~Hyde, T.~Curtois, J.~Rodr{\'{\i}}guez, J.~Walker, M.~Gendreau,
  G.~Kendall, B.~McCollum, A.~Parkes, S.~Petrovic, and E.~Burke.
\newblock Hyflex: {A} benchmark framework for cross-domain heuristic search.
\newblock In {\em Proc. of EvoCOP}, pages 136--147, 2012.

\bibitem[\protect\citeauthoryear{Senior \bgroup \em et al.\egroup
  }{2013}]{senior-icassp13}
A.~Senior, G.~Heigold, M.~Ranzato, and K.~Yang.
\newblock An empirical study of learning rates in deep neural networks for
  speech recognition.
\newblock In {\em Proc. of {ICASSP}}, 2013.

\bibitem[\protect\citeauthoryear{Shahriari \bgroup \em et al.\egroup
  }{2016}]{shahriari-ieee16a}
B.~Shahriari, K.~Swersky, Z.~Wang, R.~Adams, and N.~de~Freitas.
\newblock Taking the human out of the loop: {A} review of {B}ayesian
  optimization.
\newblock {\em Proceedings of the {IEEE}}, 104(1):148--175, 2016.

\bibitem[\protect\citeauthoryear{Shala \bgroup \em et al.\egroup
  }{2020}]{shala-ppsn20}
G.~Shala, A.~Biedenkapp, N.~Awad, S.~Adriaensen, M.~Lindauer, and F.~Hutter.
\newblock Learning step-size adaptation in {CMA-ES}.
\newblock In {\em Proc. of {PPSN}}, pages 691--706, 2020.

\bibitem[\protect\citeauthoryear{Siems \bgroup \em et al.\egroup
  }{2020}]{siems-arxiv20a}
J.~Siems, L.~Zimmer, A.~Zela, J.~Lukasik, M.~Keuper, and F.~Hutter.
\newblock {NAS-Bench-301} and the case for surrogate benchmarks for neural
  architecture search.
\newblock {\em CoRR}, abs/2008.09777, 2020.

\bibitem[\protect\citeauthoryear{Smith{-}Miles \bgroup \em et al.\egroup
  }{2014}]{smith-miles-cor14}
K.~Smith{-}Miles, D.~Baatar, B.~Wreford, and R.~Lewis.
\newblock Towards objective measures of algorithm performance across instance
  space.
\newblock {\em Comput. Oper. Res.}, 45:12--24, 2014.

\bibitem[\protect\citeauthoryear{Speck \bgroup \em et al.\egroup
  }{2020}]{speck-prl20}
D.~Speck, A.~Biedenkapp, F.~Hutter, R.~Mattm\"uller, and M.~Lindauer.
\newblock Learning heuristic selection with dynamic algorithm configuration.
\newblock In {\em Workshop on Bridging the Gap Between AI Planning and
  Reinforcement Learning ({PRL}@{ICAPS}'20)}, October 2020.

\bibitem[\protect\citeauthoryear{Speck \bgroup \em et al.\egroup
  }{2021}]{speck-icaps21}
D.~Speck, A.~Biedenkapp, F.~Hutter, R.~Mattm\"uller, and M.~Lindauer.
\newblock Learning heuristic selection with dynamic algorithm configuration.
\newblock In {\em Proc. of {ICAPS}'21}, August 2021.

\bibitem[\protect\citeauthoryear{Vermetten \bgroup \em et al.\egroup
  }{2019}]{vermetten-gecco19}
D.~Vermetten, S.~van Rijn, T.~B{\"{a}}ck, and C.~Doerr.
\newblock Online selection of {CMA-ES} variants.
\newblock In {\em Proc. of {GECCO}}, pages 951--959. {ACM}, 2019.

\bibitem[\protect\citeauthoryear{Ying \bgroup \em et al.\egroup
  }{2019}]{ying-icml19a}
C.~Ying, A.~Klein, E.~Christiansen, E.~Real, K.~Murphy, and F.~Hutter.
\newblock Nas-bench-101: Towards reproducible neural architecture search.
\newblock In {\em Proc. of {ICML}}, pages 7105--7114, 2019.

\end{thebibliography}

\appendix
\renewcommand{\thetable}{\Alph{section}\arabic{table}}
\renewcommand{\thefigure}{\Alph{section}\arabic{figure}}
\setcounter{figure}{0}

\section{Structure \& Implementation}\label{app:implementation}
Section~\ref{considerations} of the main paper discussed necessary design decisions of a DAC benchmark. Here we describe in greater detail how we chose to implement these ideas in practice.

\paragraph{Structure} A crucial first decision that shaped the project is the structure of DACBench.
It is designed to enable researchers to modify benchmarks more easily than currently possible, but also to replicate diverse experiment settings quickly.
To accomplish this, we initialize the benchmark environments with an experiment configuration that can be stored and shared easily.
Researchers can then load this configuration and reproduce the experiment (see also Section~\ref{sec:mod}).
The benchmark configuration focuses on the specification options for mentioned in Section~\ref{components} of the main paper.

\paragraph{Algorithm Implementation}
We provide or specify the implementation as well the definition of what constitutes a \emph{step} in this implementation in order to keep execution consistent. 
As researchers may not focus on dynamically optimizing all hyperparameters in parallel, the hyperparameters of the target algorithm that are not currently controlled can be modified prior and are kept fixed throughout a run. 
This can also include utility parameters concerning the DAC interface, parallelization or using different execution modes of the target algorithm.

\paragraph{Action Space}
Here the user can specify the search space for the hyperparameters being controlled.
It is important that a step of the target-algorithm knows how to parse the parameter values.
Otherwise, the order of parameters might get mixed and can lead to faulty behaviour.

\paragraph{State \& Reward}
Researchers can use the default methods for computing state or reward, but they can also provide custom ones that better fit their needs. 
DACBench can record and reconstruct experiments with custom functions as long as their implementation is shared with the experiment description.

\paragraph{Instance Sets}
Lastly, we provide instance sets for training and testing, either directly from published versions of the benchmarks or sampled from distributions that are as similar as possible.
The training and tests sets are drawn from the same distribution and contain $100$ instances each.
We believe using fixed instance sets will make comparisons between methods more consistent overall.
Nevertheless, of course different instance sets as well instances sampled on the fly can be used to better explore how DAC methods behave in more general cases or specific regions of the instance space.
While we strive to record experiment specifications as succinct as possible, the instances themselves will have to be shared by the researchers working with them.
Additionally users can modify the experimental setup including the seed, utilities for logging results and other relevant information like additional functionality like added noise.
These are often important for reproducibility and should be shared with other researchers along with the benchmark specification itself.

\paragraph{Interface}The interface we use for conducting these experiments is heavily based on the OpenAI gym~\cite{gym} interface for its simplicity.
Users only need to initialize the environment and the call two methods, reset and step, to interact with it.
Furthermore, it provides an easy way to specify action and state spaces.
As this the most commonly used RL interface, it will be very easy for RL experts to work on DAC, making the field as a whole more approachable.
To ensure proper initialization using the configuration, environments in DACBench are created by benchmark classes, each holding a default configuration that can be expanded upon and saved.
Additionally, we provide an evaluation method for the whole benchmark suite, to make comparisons to baselines and other methods simple and consistent.
We provide data of standard baselines like random policies to standardize these comparisons further.
As a lot of the benchmark specification is done through the configuration, smaller changes like the use of a different state function can be done by domain experts without modifying the environment code.
We believe this will benefit the field in the long run as experts can improve benchmarks with little effort and without working through all of the existing code base.
This will lead to better benchmarks and thriving community around DAC.

\paragraph{Additional functionality}
We provide functionality to add randomness to the benchmarks in the form of sampled instances from custom distributions as well as noise on the reward signal.
We do not include these in the default benchmarks to keep execution consistent but they provide important insight in algorithm behaviour nonetheless.
Moreover, we include smaller benchmark suites that provide both research direction and progress measure in a very efficient format. 
Lastly, DACBench also includes a flexible logging system that allows the user to track performance data as well as action trajectories, state features and execution times, all sorted by environment step and instance used.
To make analysis quick and convenient, we include methods for reloading data as well as visualization tools.

\section{The Benchmarks in detail}\label{app:benchmarks}
\paragraph{Sigmoid} The Sigmoid benchmark \citep{biedenkapp-ecai20} is parameterized by the number of functions to approximate in parallel in the number of actions in each of these dimensions (by default two dimensions with $10$ and $5$ actions respectively ). 
An instance provides shift and slope for the sigmoid function in each of these dimensions such that the function value is:
\begin{equation*}
    sigmoid(t) = \frac{1}{1 + e^{-slope * (t-shift)}}
\end{equation*}
The reward in each dimension is then the distance from the chosen action to the function value. 
The rewards from each dimension are then multiplied to form the step reward.
The state is made up of the current step, previous state (including all slopes and shifts for the current instance), action played, reward and next state.

\paragraph{Luby} This benchmark has the agent approximate the Luby sequence of a given length.
(Following \cite{biedenkapp-ecai20} the length is pre-set to $64$, resulting in $6$ distinct actions). 
In the default setting, this is challenging enough, but instances for Luby can be used to modify the start value of the sequence or add an accumulating error signal. 
The reward is $0$ if the correct action is played, $-1$ otherwise. 
The observations returned consist of the current timestep, previous state (including a history of actions for the last $5$ steps), action played, reward, next state, next goal and time cutoff.

\paragraph{FastDownward} In FastDownward, the agent selects a search heuristic for the next planning interval.
The default action space consist of two heuristics, although we also provide a more complete version with four heuristics \citep{speck-prl20}.
The two heuristic version is an easier variation with artificial instances that provides an easier optimization variation and it has been shown that an RL agent is capable of recovering the optimal policy on these \citep{speck-prl20}.
The reward is $-1$ per step, measuring the total number of steps.
The agent observes average, minimum and maximum values as well as variance and open list entries for each heuristic.
We provide several target domains as instance sets with the default being an artificially generated one.
 
\paragraph{CMA-ES} Following \cite{shala-ppsn20}, an agent's task here is to adjust the step size for CMA-ES. 
Therefore the actions space covers possible step sizes between $0$ and $10$.
The reward is the best individual's objective value, with the observations consisting of the current step size, cumulative path length and population size as well as past objective values, change in past objective values and past step sizes (for the last $40$ steps each).
The instance set consists of $10$ different function classes of the BBOB benchmark.
 
\paragraph{ModEA} Instead of controlling a single hyperparameter, in ModEA the algorithm structure consisting of $11$ different components is adjusted \citep{vermetten-gecco19}. 
There are two choices for each of the first $nine$ of these and three for the other two.
That results in $4,608$ possible actions in total.
The reward is, as in CMA-ES, the best individual's fitness.
The state contains the generation size, current step size, remaining budget, function ID and instance ID.
As the large action space makes this benchmark hard, we used the same instance set as for CMA-ES with only 10 different function classes.

\paragraph{SGD-DL} Here the agent controls a small neural network's learning rate \cite{daniel-aaai16}. Instances  consist of a dataset, seed and network. In the default setting, we consider a single dataset (MNIST), $100$ seeds each for training and test, affecting weight initialization and the mini-batch ordering, and a single fully-connected feedforward network architecture having two hidden layers with $16$ units each. Permitted actions lie between $0$ and $10$ with the learning rate of the optimizer (Adam) being set to $10^{-action}$.
The observations includes discounted average and uncertainty each of the predictive change variance and loss variance, the current learning rate, the training and validation loss.

\section{Modifiers of Benchmarks}
\label{sec:mod}

\begin{figure}[tbhp]
    \centering
    \includegraphics[scale=0.4]{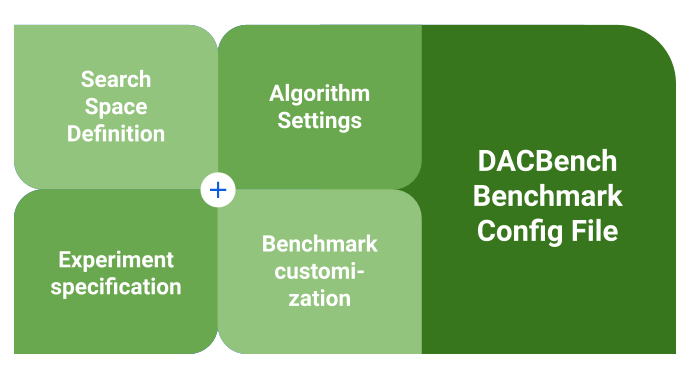}
    \caption{Modification possibilities.} %    \label{fig:benchmark_customization}
    \label{fig:mods}
\end{figure}
DACBench not only allows to use existing benchmarks, but also enables easy modification through the use of Benchmark Configurations themselves (see Figure~\ref{fig:mods}).

Modifications to the search space definition include changes to the state and action spaces.
Such changes can immediately increase or lower the difficulty of learning successful policies, e.g. through inclusion/exclusion of irrelevant action choices or use of  additional (un-)informative state features.
Changing the algorithm settings include changes to hyperparameters that are not dynamically changed or different resource allocation.
Further benchmark customization could allow control of multiple parameters (if not already enabled) or use of different problem instances.
All included benchmarks come with default configurations that allow for reproduction of experiments from published versions of the benchmarks.

\section{Experiment Hardware}\label{app:hardware}
All experiments with were conducted on a slurm CPU cluster (see Table \ref{tab:cpu_cluster}). 
The upper memory limit for these experiments was 800MB per run, although not all benchmarks require this much.
As the DACBench runner was used for all experiments, the provided Slurm scripts can reproduce the results.
Additionally, we provide them with the code.

\begin{table}[ht]
    \centering
    \begin{tabular}{c|c|c|c}
         Machine no. &  CPU model & cores & RAM \\
         \hline
          1 &  Xeon E5-2670 & 16 & 188 GB \\
          2 & Xeon E5-2680 v3 & 24 & 251 \\
          3-6 & Xeon E5-2690 v2 & 20 & 125 GB \\
          7-10 &  Xeon Gold 5120 & 28 & 187 \\
    \end{tabular}
    \caption{CPU cluster used for experiments.}
    \label{tab:cpu_cluster}
\end{table}

\section{On Quantifying the Challenge Dimensions}\label{app:dimension}
We quantify all of our dimensions of difficulty for better comparability. 
This section details our criteria.

\paragraph{State \& Action Spaces}First, state and action space size are deciding factors in MDPs. If state or action spaces are larger, learning will take longer and the probability of finding a local optimum instead of the best solution could increase for many methods.
Therefore a small set of possible states and actions makes a benchmark easier to solve regardless of how complex the underlying function is.
To make comparison between the other aspects as well as discrete and continuous action easy, we divide the spaces into three categories.
Category one contains small discrete action spaces, we define this as below $100$ actions. Large discrete spaces with up to $1000$ actions fall into category two. For state spaces, this means spaces of up to $100$ dimensions. Larger spaces fall into category three. Continuous action spaces and action spaces with more than $1000$ possible actions fall into the same category.
This is of course only a very rough categorization, but it should provide an overview of how our benchmarks differ.

\paragraph{Reward Function}In building DAC benchmarks, deciding on a reward signal is as important as it can be difficult. 
A good reward signal would attribute every action a reward proportional to its quality. 
This is hard to accomplish and sometimes we have to default to a very sparse reward signal, requiring the agent to learn to interpret the reward. 
A more informative reward, what we call better reward quality, is therefore a desirable quality from a learning perspective.
For this category, we define a scale from $1$ and $5$: $5$ means no meaningful reward at all, $4$ is a combined reward only at the end with no information (reward of $0$ in each step) during the run. A score of $3$ is similar, a meaningful signal only at the end but with step rewards that indicate if taking steps is desired or not (e.g. now giving $+1$ or $-1$ per step). A reward of quality $2$ provides the accumulated quality of the policy so far at each step, but not how the last action contributed specifically. Lastly, if the reward indicates directly how good of a choice the last action specifically was, it would be of quality level $1$.
Just like the action and state space size, we can judge this benchmark feature without any empirical evaluations.

\paragraph{Noise}As DAC means working with algorithms that may not have exact same execution times and patterns across different runs and hardware setups, a DAC agent should be able to learn and perform in noisy settings. 
Therefore we consider reward noise, which makes finding the target policy harder, an important challenge in DAC. 
To measure it, we need to run the benchmarks. We chose random policies and compute the standard deviation normalized by the mean between $10$ evaluations of the baseline policies per seed and average over $10$ seeds. 
\paragraph{Policy heterogeneity}Policy heterogeneity is another component of benchmark difficulty. 
If policies across benchmark instances stay relatively similar, they should be easier and faster to learn, as all instances provide the same or at least a similar signal for optimizing the policy.
As we do not have access to the optimal policy for our benchmarks, we use the results of our static policy evaluation as an estimation of how well a single policy can cover the instance space.
We compare the average standard deviation normalized by the mean static policies show across all instances.

\paragraph{Dynamicity}Lastly, we examine how dynamic our benchmarks are, that is how many action changes we expect in a policy. If a benchmark is not very dynamic, policies that only update the hyperparameter value once or twice might be best while highly dynamic benchmarks require almost constant changes.
Again we lack the optimal policies to get a definite answer to this question, but we approximate it using static and random policies.
For each benchmark, we cover the given search space with a number of static policies and run random policies with repeating actions. Actions are repeated for a total of $1$, $10$, $100$ or $1000$ steps. As before, we evaluate each repeat number for $10$ seeds with $10$ runs each. 
The benchmarks are scored depending on the performance ranks of these random policies. If the policy with only $1$ repetition performs best on average on an instance per seed, its score is increased by $3$. $10$ repetitions yield $2$ points, $100$ $1$ point and no points if the policy with $1000$ repeats performs best. 
We then scale this score to $(0, 1)$, as we do with all others, for simplicity's sake.

\section{Additional experimental results}
\setcounter{figure}{0}
For space reasons, we did not include all comparisons of static and dynamic policies in the main paper.
As it is still interesting to see how dynamic baselines and random policies compare to the best and worst static policies, we include the missing benchmarks here.
Within these, we can easily identify two groups: on the Sigmoid and FastDownward benchmarks, dynamic policies obviously perform well.
FastDownward in particular seems to favor heavily dynamic policies, as our previous analysis has shown already.
For ModEA and CMA-ES, the picture is not quite so clear, with the random policy and even CSA for CMA-ES peforming somewhere in between the best and worst static policies.
As we have seen results on these problem settings that suggest good dynamic policies perform far better, however, we can simply assume CMA-ES and ModEA to be harder benchmarks to beat static policies on.
As they both have large action spaces as well as high policy heterogeneity and noise rating in addition to this fact, they present the upper end of difficulty in DACBench.

\begin{figure}[tbh]
    \centering
    \includegraphics[scale=0.25]{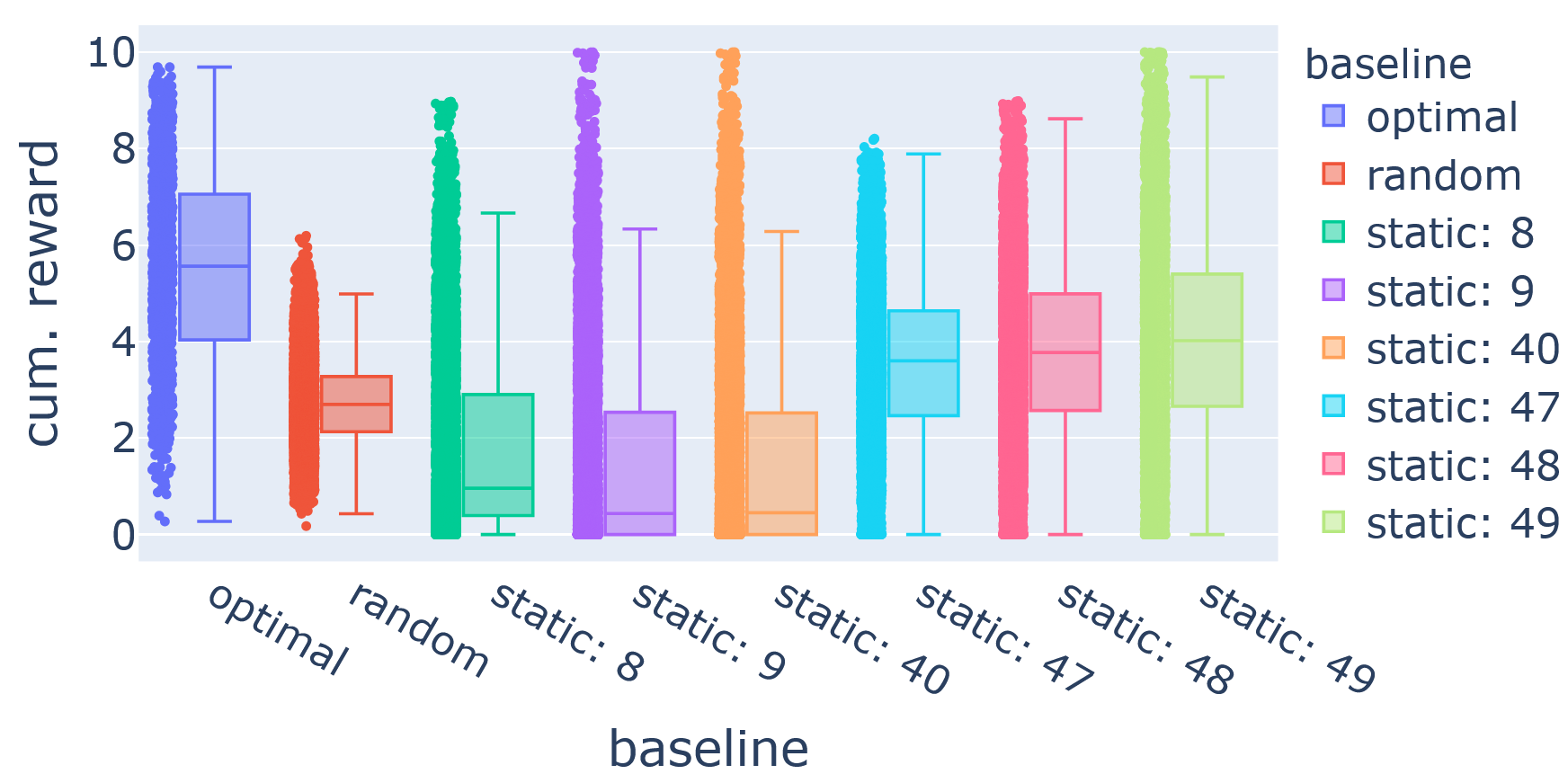}
    \caption{Static and dynamic policies on Sigmoid including top \& bottom 3 static policies. The reward measures how close the chosen discrete value is to the actual funvtion value.}
    \label{app-fig:policies_sigmoid}
\end{figure}

\begin{figure}[tbh]
    \centering
    \includegraphics[scale=0.25]{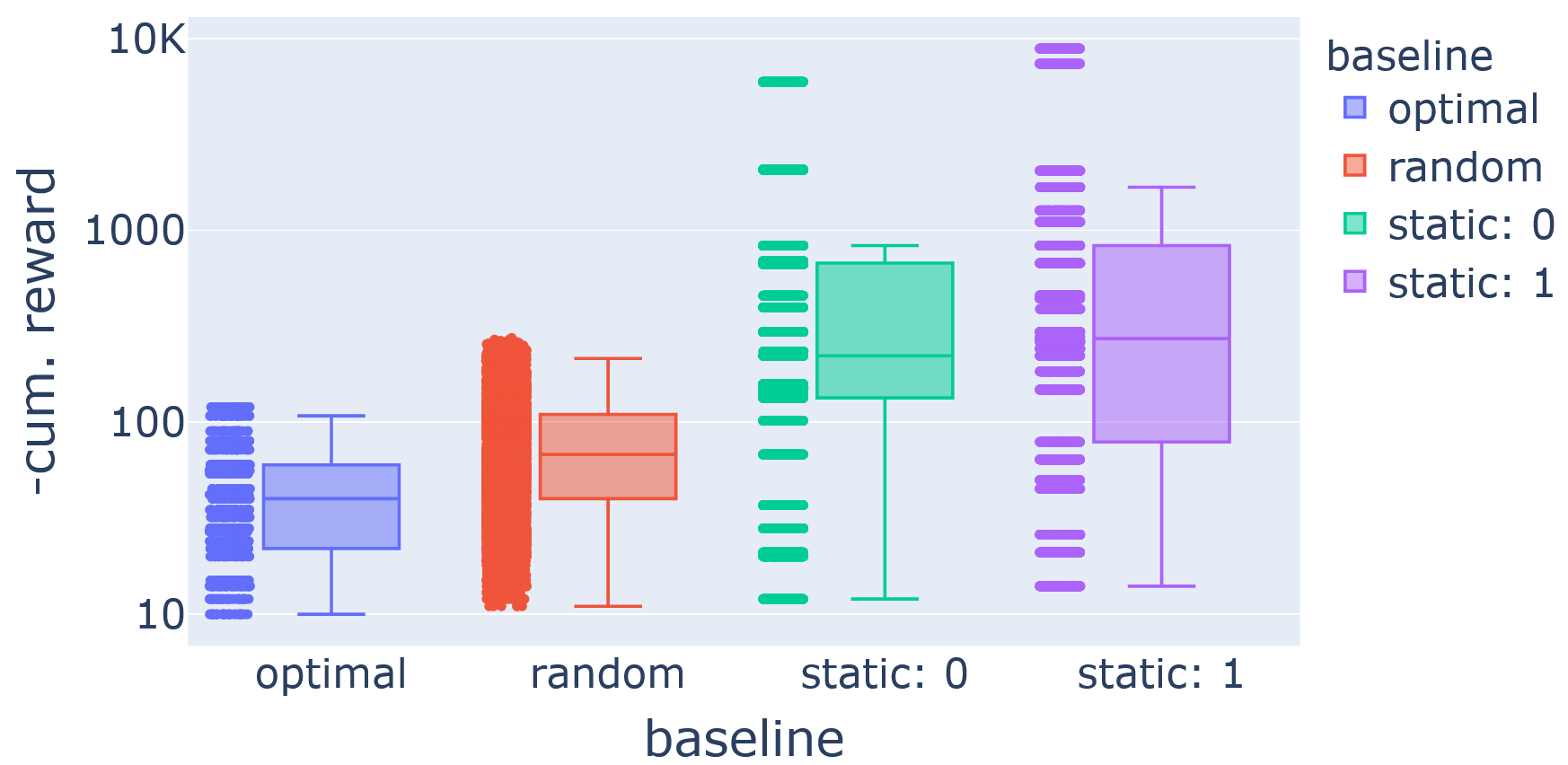}
    \caption{Static and dynamic policies on FastDownward. The reward is $-1$ per step until the run is finished.}
    \label{fig:policies_fd}
\end{figure}

\begin{figure}[tbh]
    \centering
    \includegraphics[scale=0.25]{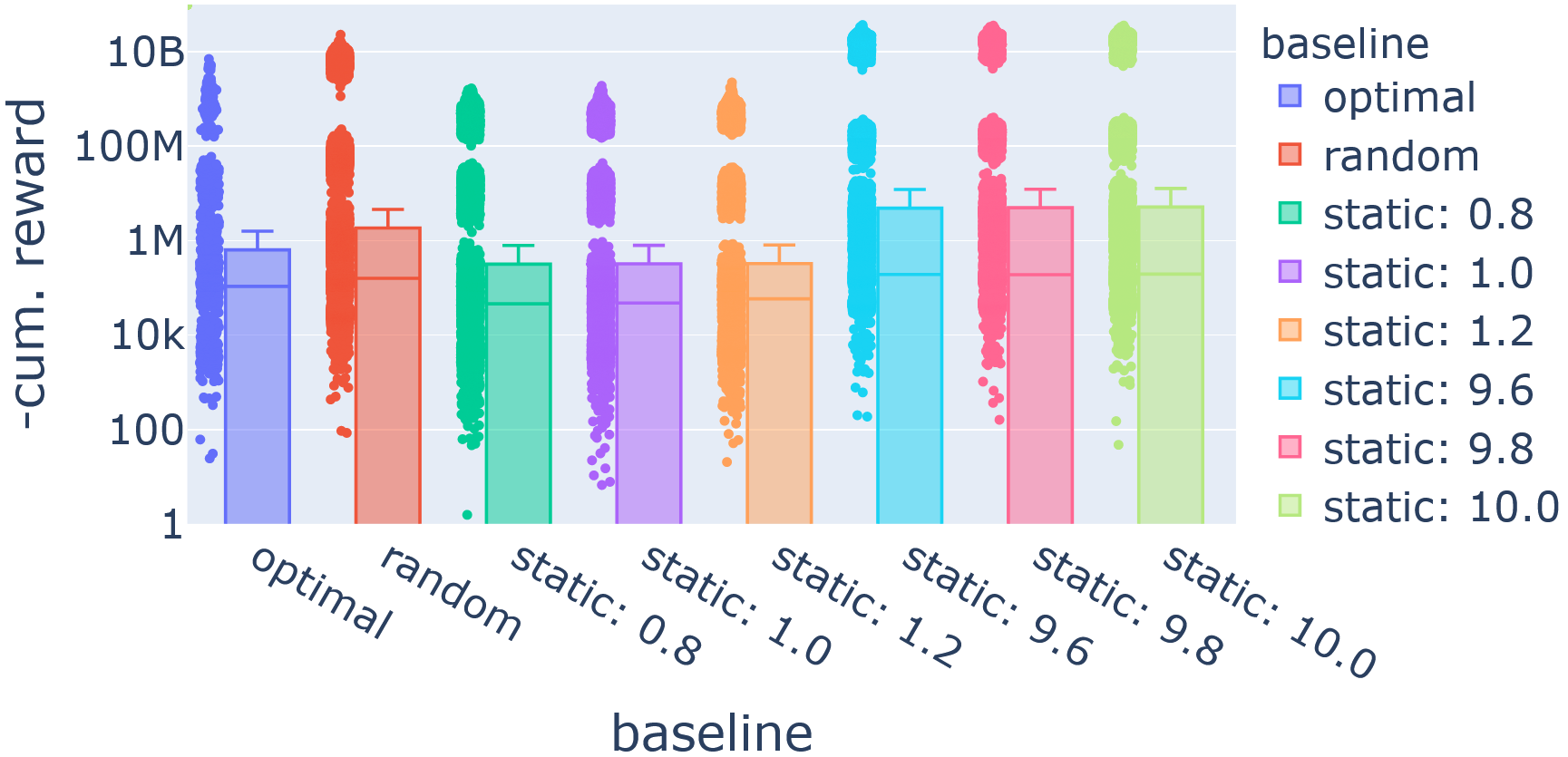}
    \caption{Static and dynamic policies on CMA-ES, including top \& bottom 3 policies and CSA. The reward is the fitness of the best indivudual in the population.}
    \label{app-fig:policies_cma}
\end{figure}

\begin{figure}[tbh]
    \centering
    \includegraphics[scale=0.25]{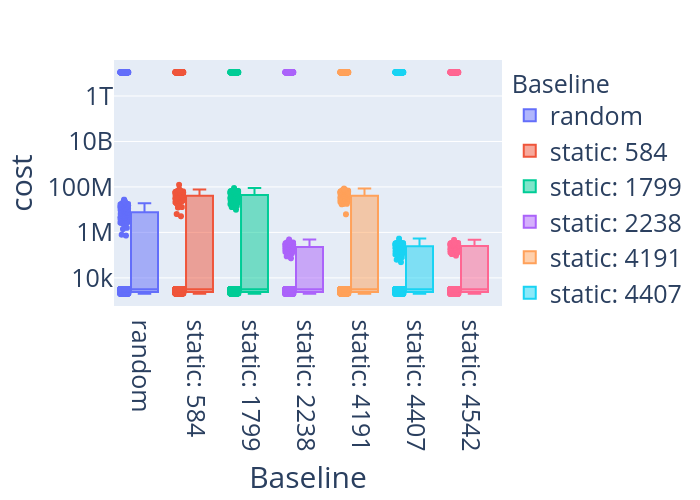}
    \caption{Static and dynamic policies on ModEA, including top \& bottom 3 policies. The reward, as in CMA-ES, is the current best individual.}
    \label{fig:policies_modea}
\end{figure}
\end{document}